\documentclass[conference]{IEEEtran}
\IEEEoverridecommandlockouts
\usepackage[T5]{fontenc}
\usepackage[utf8]{inputenc}
\usepackage{cite}
\usepackage{amsmath,amssymb,amsfonts}
\usepackage{algorithmic}
\usepackage{graphicx}
\usepackage{textcomp}
\usepackage{xcolor}
\usepackage{multirow}
\usepackage{float}
\usepackage{booktabs}
\usepackage{comment}
\usepackage[caption=false,font=footnotesize]{subfig}
\usepackage{url}

\usepackage{tikz}
\usetikzlibrary{shapes.geometric,arrows.meta,positioning,fit,calc,backgrounds,decorations.pathreplacing}
\usepackage[colorlinks=true,citecolor=red]{hyperref}
\def\BibTeX{{\rm B\kern-.05em{\sc i\kern-.025em b}\kern-.08em
    T\kern-.1667em\lower.7ex\hbox{E}\kern-.125emX}}

\begin{document}

\title{A Hardware-Aware Comprehensive Study of Edge Vision Deployment with A Unified Hardware-Model-Runtime Mapping Framework}

% \author{\IEEEauthorblockN{Le Duc An}
% \IEEEauthorblockA{\textit{Faculty of Information Technology} \\
% \textit{University of Science, VNUHCM}\\
% Ho Chi Minh City, Viet Nam \\
% 23120109@student.hcmus.edu.vn}
% \and
% \IEEEauthorblockN{Le Duc Thanh}
% \IEEEauthorblockA{\textit{Faculty of Information Technology} \\
% \textit{University of Science, VNUHCM}\\
% Ho Chi Minh City, Viet Nam \\
% 23120165@student.hcmus.edu.vn}
% }

\author{
\IEEEauthorblockN{
Le Duc An\textsuperscript{a,b},
Le Duc Thanh\textsuperscript{a,b},
Vo Hoai Viet\textsuperscript{a,b}
}
\IEEEauthorblockA{
\textsuperscript{a}\textit{Faculty of Information Technology, University of Science, VNU-HCM}\\
\textsuperscript{b}\textit{Vietnam National University, Ho Chi Minh City, Vietnam}
}
\thanks{Corresponding author: Vo Hoai Viet (email: vhviet@fit.hcmus.edu.vn)}
}

\maketitle

\begin{abstract}
The rapid growth of deep learning has driven the deployment of computer vision directly on resource-constrained edge devices. However, the extreme heterogeneity of edge hardware architectures poses major challenges in selecting and designing suitable vision models. This survey bridges the gap between edge hardware constraints, model architectures, and deployment software toolchains. We establish a comprehensive taxonomy classifying edge hardware into three distinct tiers: mid-range devices (GPUs and dedicated NPUs/TPUs), dedicated embedded MPU/SoC processors, and microcontrollers (MCUs). We aggregate and analyze multi-source benchmark results for image classification and object detection across the computing spectrum, spanning from tens of watts down to milliwatt-scale bare-metal MCUs. Based on these empirical findings, we propose a two-tiered hardware-to-model mapping framework that provides concrete model selection matrices across performance priorities (Speed, Balanced, Accuracy) and actionable architectural design principles (peak SRAM budgeting, resolution-first scaling, and algorithm--compiler co-design). Finally, we highlight critical deployment challenges—including toolchain fragmentation, operator incompatibility, and Vision Transformer feasibility—and outline promising future research directions in on-device adaptation and sustainable Edge AI.

\end{abstract}

\begin{IEEEkeywords}
Hardware-Model-Runtime Mapping, Computer Vision, Edge Vision, Edge AI, IoT.
\end{IEEEkeywords}

\section{Introduction}\label{sec:intro}
Over the past decade, deep learning models in computer vision have achieved remarkable accuracy across fundamental tasks such as image classification, object detection, and semantic segmentation, driven by increasingly deep convolutional neural networks and attention-based architectures. In parallel with algorithmic advancements, a critical paradigm shift is underway in real-world deployments: transferring computer vision inference directly from centralized cloud data centers to resource-constrained edge devices located at the data source. This transition is motivated by four stringent operational imperatives: (1) deterministic low latency and real-time responsiveness, (2) robust operation in network-disconnected or intermittent environments, (3) elimination of recurring cloud bandwidth and ingress costs for continuous video streams, and (4) strict privacy and compliance guarantees achieved by processing sensitive visual data on-device without cloud transmission. Representative applications driving this shift include autonomous robotics, industrial quality inspection, smart agricultural monitoring, and camera-equipped Internet of Things (IoT) endpoints.

A defining characteristic of edge vision deployment compared to cloud environments is the extreme architectural heterogeneity of target hardware. The edge computing spectrum spans from mid-range embedded platforms equipped with dedicated GPUs or neural accelerators (such as NVIDIA Jetson, Google Coral, and Hailo AI accelerator)—capable of executing multi-million-parameter models at hundreds of frames per second—down to ultra-low-power microcontrollers (MCUs) constrained by sub-megabyte SRAM and strict milliwatt power budgets. In this low-power domain, the TinyML and TinyDL paradigms have emerged to bring deep intelligence directly to sensor nodes \cite{Somvanshi2025FromTM, lin2023tiny}. On every edge platform, the triad of physical hardware limits—compute capacity, memory hierarchy, and power envelope—strictly dictates feasible model topologies, while simultaneously determining compatible runtime compilers and precision formats.

Recent literature has advanced edge vision through three prominent thrusts \cite{Somvanshi2025FromTM}: (i) lightweight architectural innovations, such as depthwise separable convolutions, inverted residual bottlenecks, and compact attention mechanisms; (ii) model compression techniques, including post-training quantization (PTQ), structured pruning, and knowledge distillation; and (iii) hardware-aware Neural Architecture Search (NAS) frameworks that search for Pareto-optimal network topologies under physical constraints. However, existing surveys frequently present model compression and architectural scaling as abstract, hardware-agnostic toolkits without explicitly linking them to the quantitative resource budgets (SRAM boundaries, peak TOPS, memory bandwidth, and compiler support) of specific edge hardware tiers.

\begin{table*}[t]
\centering
\caption{Comparison of Representative Edge AI and Vision Deployment Surveys with Our Work}
\label{tab:related_surveys}
\resizebox{\textwidth}{!}{
\begin{tabular}{lccccccc}
\toprule
\textbf{Survey Reference} & \textbf{Year} & \textbf{Hardware Scope} & \textbf{Vision Focus} & \textbf{Runtime Ecosystem} & \textbf{Cross-Tier Benchmarks} & \textbf{MCU / TinyML} & \textbf{HW $\to$ Model Mapping} \\
\midrule
Saha \textit{et al.}~\cite{saha2022machine}                       & 2022 & MCUs only                   & General ML               & MCU runtimes (TFLM)            & Limited                     & \checkmark & $\times$ \\
Lin \textit{et al.}~\cite{lin2023tiny}                            & 2023 & MCUs only                   & General TinyML           & TinyEngine / TFLM              & Limited                     & \checkmark & Partial \\
Iqbal \textit{et al.}~\cite{Iqbal2024ARO}                         & 2024 & Mid-range GPU SBCs          & Vision (CNNs, Detectors) & GPU toolchains (TensorRT)      & Single-tier (SBCs)          & $\times$   & $\times$ \\
Papa \textit{et al.}~\cite{papa2024survey}                        & 2024 & Generic Edge / GPUs         & Vision (ViTs only)       & N/A (Algorithmic)              & Framework-level             & $\times$   & $\times$ \\
Wang \& Jia~\cite{Wang2025OptimizingEA}                           & 2025 & Generic Edge / Cloud        & General Edge AI          & High-level frameworks          & $\times$                    & Partial    & $\times$ \\
Somvanshi \textit{et al.}~\cite{Somvanshi2025FromTM}              & 2025 & MCUs \& Embedded Nodes      & General TinyDL           & Micro-runtimes                 & Qualitative                 & \checkmark & $\times$ \\
Saha \& Xu~\cite{Saha2025VisionTO}                                & 2025 & Edge NPUs / GPUs            & Vision (ViTs only)       & Specialized accelerators       & Model-level only            & $\times$   & $\times$ \\
Ali \& Dornaika~\cite{ali2025edge}                                & 2025 & Conceptual Edge Taxonomy    & General Edge AI          & Generic toolchains             & $\times$                    & Partial    & $\times$ \\
Cordova-Cardenas \textit{et al.}~\cite{CordovaCardenas2025EdgeAI} & 2025 & Embedded SoCs               & General Deep Learning    & Embedded runtimes              & Limited                     & $\times$   & Partial \\
Ratul \textit{et al.}~\cite{ratul2025accelerating}                & 2025 & Edge GPUs / SoCs (AGX Orin) & General DL Inference     & \checkmark (Inference engines) & Framework-level             & $\times$   & $\times$ \\
\midrule
\textbf{Ours} & \textbf{2026} & \textbf{Full Spectrum (GPU/NPU/MPU/MCU)} & \textbf{\checkmark (CNNs, Detectors, ViTs)} & \textbf{\checkmark (Mid-range to MCU engines)} & \textbf{\checkmark (Multi-Source Empirical)} & \textbf{\checkmark} & \textbf{\checkmark (Actionable Decision Matrices)} \\
\bottomrule
\end{tabular}
}
\end{table*}

As systematically summarized in Table~\ref{tab:related_surveys}, prior surveys in the literature have focused on isolated slices of the edge computing ecosystem. Hardware-centric and model-compression surveys often examine platforms or model families in isolation: Saha \& Xu~\cite{Saha2025VisionTO} and Papa et al.~\cite{papa2024survey} investigate efficiency strategies for Vision Transformers---with the former focusing on specialized hardware accelerators (FPGAs, ASICs) and hardware--software co-design, and the latter focusing primarily on algorithmic model compression (pruning, quantization, and knowledge distillation)---yet both omit the lightweight CNN families that dominate low-power MCUs. Other works evaluate single-board computers (SBCs) for real-time vision without addressing microcontroller-grade endpoints \cite{Iqbal2024ARO}. In the microcontroller domain, foundational surveys such as Saha et al.~\cite{saha2022machine}, Lin et al.~\cite{lin2023tiny}, and Somvanshi et al.~\cite{Somvanshi2025FromTM} focus primarily on TinyML algorithms, memory-efficient micro-engines (e.g., TinyEngine, TFLM), and MCU micro-architectures, but omit mid-range NPUs/GPUs, modern object detection pipelines (YOLOv8--v26), and cross-tier empirical benchmarking. Meanwhile, broader surveys explore multi-dimensional Edge AI taxonomies along generic data--model--system axes across non-visual modalities \cite{Wang2025OptimizingEA, ali2025edge}, or outline high-level experimental methodologies without synthesizing cross-platform benchmark data \cite{CordovaCardenas2025EdgeAI, ratul2025accelerating}.

Consequently, existing literature exhibits four critical research gaps: (1) an absence of a unified full-spectrum hardware taxonomy linking 25\,W accelerators down to sub-milliwatt MCUs; (2) a disconnect between vision model operators (depthwise separable blocks, linear bottlenecks, multi-scale FPNs) and low-level physical constraints (on-chip SRAM activation limits and systolic dataflows); (3) a lack of cross-tier empirical benchmark aggregation spanning classification and detection across heterogeneous devices; and (4) an absence of actionable hardware-to-model design decision matrices that translate operational priorities (Speed, Balanced, Accuracy) into concrete engineering choices.

\subsection{The 5-Level Edge Vision Deployment Framework}

\begin{figure*}[t]
\centering
\resizebox{0.98\textwidth}{!}{
\begin{tikzpicture}[
    >=Latex,
    node distance=0.25cm,
    levelbox/.style={
        rectangle,
        rounded corners=4pt,
        draw=#1,
        fill=#1!4,
        line width=0.9pt,
        text width=3.35cm,
        minimum height=3.5cm,
        align=left,
        inner sep=6pt
    },
    header/.style={
        rectangle,
        rounded corners=3pt,
        fill=#1,
        text=white,
        font=\scriptsize\bfseries,
        text width=3.35cm,
        align=center,
        inner sep=3.5pt
    }
]

% Level 1: Hardware
\node[header=blue!75!black] (h1) at (0, 1.8) {Level 1: Hardware (Sec.~II)};
\node[levelbox=blue!75!black, below=2pt of h1] (b1) {
    \begin{minipage}{3.25cm}
    \scriptsize\raggedright
    \textbf{Hardware Tiers:}\\[1.5pt]
    \mbox{\textbullet~}Mid-range GPU / NPU\\
    \mbox{\textbullet~}Dedicated MPU / SoC\\
    \mbox{\textbullet~}Microcontrollers (MCUs)\\[3pt]
    \textbf{Physical Constraints:}\\[1.5pt]
    \mbox{\textbullet~}Compute (GFLOPS/TOPS)\\
    \mbox{\textbullet~}Internal SRAM \& Flash\\
    \mbox{\textbullet~}Power (5--25\,W vs. mW)\\
    \mbox{\textbullet~}Host interconnect / bus
    \end{minipage}
};

% Level 2: Model
\node[header=teal!70!black, right=0.35cm of h1] (h2) {Level 2: Model (Sec.~III)};
\node[levelbox=teal!70!black, below=2pt of h2] (b2) {
    \begin{minipage}{3.25cm}
    \scriptsize\raggedright
    \textbf{Vision Models:}\\[1.5pt]
    \mbox{\textbullet~}Lightweight CNNs\\
    \mbox{\textbullet~}Baseline CNNs (ResNet)\\
    \mbox{\textbullet~}Detectors (YOLO, SSD)\\
    \mbox{\textbullet~}Lightweight / Hybrid ViTs\\[3pt]
    \textbf{Scaling \& Operators:}\\[1.5pt]
    \mbox{\textbullet~}Resolution scale ($\rho$)\\
    \mbox{\textbullet~}Width / Depth ($\alpha, d$)\\
    \mbox{\textbullet~}Inverted residuals / DW\\
    \mbox{\textbullet~}INT8 Quantization / NAS
    \end{minipage}
};

% Level 3: Runtime
\node[header=orange!85!black, right=0.35cm of h2] (h3) {Level 3: Runtime (Sec.~IV)};
\node[levelbox=orange!85!black, below=2pt of h3] (b3) {
    \begin{minipage}{3.25cm}
    \scriptsize\raggedright
    \textbf{Deployment Engines:}\\[1.5pt]
    \mbox{\textbullet~}TensorRT, Edge TPU\\
    \mbox{\textbullet~}HailoRT, OpenVINO\\
    \mbox{\textbullet~}TFLM, CMSIS-NN\\
    \mbox{\textbullet~}TinyEngine, TinyOps, SDKs\\[3pt]
    \textbf{Compilation \& Tuning:}\\[1.5pt]
    \mbox{\textbullet~}Kernel fusion \& compile\\
    \mbox{\textbullet~}Static memory layout\\
    \mbox{\textbullet~}Operator fallback routing
    \end{minipage}
};

% Level 4: Benchmarks
\node[header=purple!75!black, right=0.35cm of h3] (h4) {Level 4: Benchmark (Sec.~V)};
\node[levelbox=purple!75!black, below=2pt of h4] (b4) {
    \begin{minipage}{3.25cm}
    \scriptsize\raggedright
    \textbf{Multi-Source Evaluation:}\\[1.5pt]
    \mbox{\textbullet~}Classification (ImageNet)\\
    \mbox{\textbullet~}Detection (COCO, VOC)\\
    \mbox{\textbullet~}Cross-tier comparison\\[3pt]
    \textbf{Empirical Metrics:}\\[1.5pt]
    \mbox{\textbullet~}Throughput (FPS / Latency)\\
    \mbox{\textbullet~}Accuracy (Acc@1 / mAP)\\
    \mbox{\textbullet~}Memory (SRAM/Flash KB)\\
    \mbox{\textbullet~}Diminishing return trends
    \end{minipage}
};

% Level 5: Design Decision
\node[header=red!75!black, right=0.35cm of h4] (h5) {Level 5: Decision (Sec.~VI)};
\node[levelbox=red!75!black, below=2pt of h5] (b5) {
    \begin{minipage}{3.25cm}
    \scriptsize\raggedright
    \textbf{Deployment Priorities:}\\[1.5pt]
    \mbox{\textbullet~}Speed (Max FPS / Latency)\\
    \mbox{\textbullet~}Balanced (Pareto Trade-off)\\
    \mbox{\textbullet~}Accuracy (Max mAP)\\[3pt]
    \textbf{Decision Dimensions:}\\[1.5pt]
    \mbox{\textbullet~}Select Model: Off-the-shelf\\
    \mbox{\textbullet~}Modify Arch: SRAM budget\\
    \mbox{\textbullet~}Optimize Runtime: Co-design
    \end{minipage}
};

% Inter-box horizontal arrows
\draw[->, line width=1.1pt, gray!70!black] (b1.east |- b1.center) -- (b2.west |- b2.center);
\draw[->, line width=1.1pt, gray!70!black] (b2.east |- b2.center) -- (b3.west |- b3.center);
\draw[->, line width=1.1pt, gray!70!black] (b3.east |- b3.center) -- (b4.west |- b4.center);
\draw[->, line width=1.1pt, gray!70!black] (b4.east |- b4.center) -- (b5.west |- b5.center);

% Top-down design intent arrow (from Right Level 5 to Left Level 1)
\draw[->, line width=0.9pt, gray!85!black, dashed] ($(h5.north east)+(0,0.18)$) -- ($(h1.north west)+(0,0.18)$)
    node[midway, above, font=\scriptsize\bfseries, text=black!90] {Top-Down Deployment Requirements (Application SLA, Accuracy Target, Latency Budget)};

% Bottom-up constraint arrow (from Left Level 1 to Right Level 5)
\draw[->, line width=0.9pt, gray!85!black, dashed] ($(b1.south west)+(0,-0.18)$) -- ($(b5.south east)+(0,-0.18)$)
    node[midway, below, font=\scriptsize\bfseries, text=black!90] {Bottom-Up Physical Constraints (SRAM, Flash, TOPS, Memory Wall, Host I/O)};

\end{tikzpicture}
}
\caption{The proposed 5-Level Edge Vision Deployment Framework: Level 1 (Hardware Tier \& Constraints), Level 2 (Vision Model Space \& Scaling Knobs), Level 3 (Deployment Software Ecosystem \& Runtimes), Level 4 (Empirical Benchmarks \& Multi-Source Evaluation), and Level 5 (Actionable Design Decisions: Priority Regimes, Model Selection, Architectural Customization, and Runtime Optimization).}
\label{fig:framework_overview}
\end{figure*}
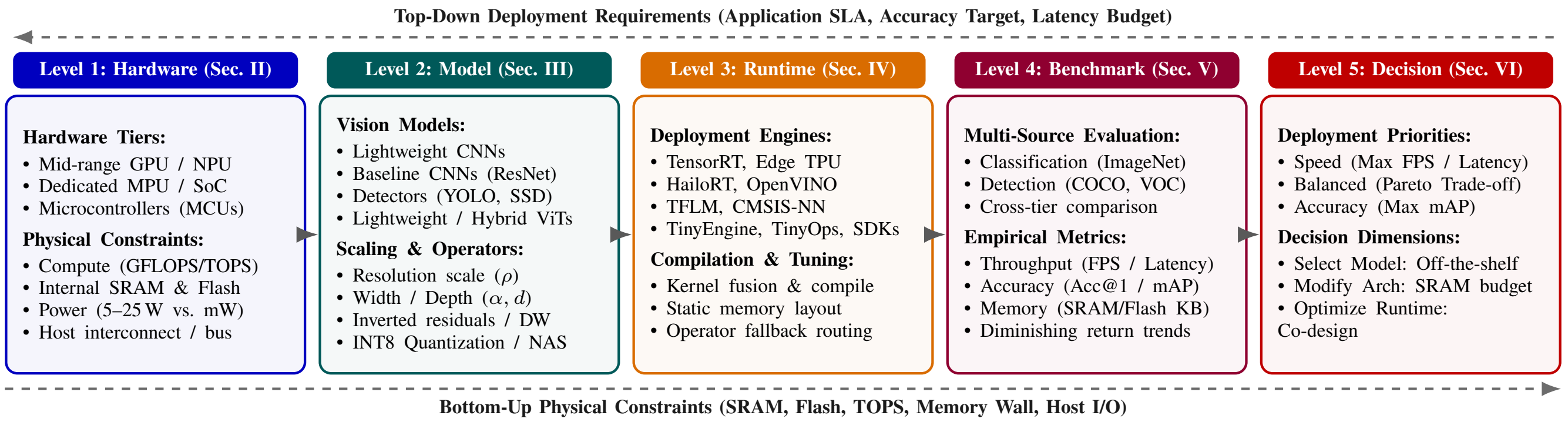

To bridge these fundamental gaps, this survey introduces a unified \textbf{5-Level Edge Vision Deployment Framework}, illustrated in Fig.~\ref{fig:framework_overview}. This framework formalizes the vertical stack governing edge vision deployment:

\begin{enumerate}
    \item \textbf{Level 1 --- Hardware Tier and Resource Constraints (Section~\ref{sec:hardware}):} Categorizes edge hardware into three distinct tiers: \textit{Mid-Range Devices} (GPUs, NPUs, TPUs), \textit{Dedicated Embedded MPUs/SoCs}, and \textit{Microcontrollers (MCUs)}. It analyzes their physical boundaries across compute capacity (GFLOPS/TOPS), memory hierarchy (on-chip SRAM, Flash, external DRAM), power envelopes (5--25\,W vs. mW scale), and host-accelerator interconnects.
    \item \textbf{Level 2 --- Vision Model Space and Scaling Knobs (Section~\ref{sec:model}):} Analyzes lightweight CNN backbones (MobileNetV1--V3, MnasNet, ProxylessNAS, MCUNet) alongside standard baseline architectures (Inception, ResNet) evaluated in benchmarks, one-stage object detectors (YOLOv3--v26, SSD-MobileNet, MobileDet, EfficientDet-Lite), and a brief synthesis of lightweight and hybrid Vision Transformers. It examines key architectural scaling knobs: resolution multiplier ($\rho$), width multiplier ($\alpha$), depth scaling ($d$), and specialized operators (depthwise separable blocks, linear bottlenecks, and spatial attention).
    \item \textbf{Level 3 --- Deployment Software Ecosystem and Runtimes (Section~\ref{sec:software}):} Investigates the compiler and execution runtime stack across tiers, including mid-range toolchains (NVIDIA TensorRT, Google Edge TPU Compiler, HailoRT, Intel OpenVINO), standard microcontroller engines (TensorFlow Lite for Microcontrollers, ARM CMSIS-NN, STMicroelectronics STM32Cube.AI), and memory-optimized academic engines (TinyEngine, TinyOps).
    \item \textbf{Level 4 --- Empirical Benchmarks and Performance Analysis (Section~\ref{sec:benchmark}):} Synthesizes multi-source empirical performance data for image classification and object detection across hardware platforms, evaluating throughput (FPS and latency in milliseconds), accuracy (Top-1 accuracy and mAP), memory footprints (SRAM and Flash budgets), and diminishing returns across scaling regimes.
    \item \textbf{Level 5 --- Actionable Design Decisions (Section~\ref{sec:framework}):} Establishes the decision-making framework translating empirical performance and application priorities (\textit{Speed Priority}, \textit{Balanced Priority}, \textit{Accuracy Priority}) into three concrete engineering dimensions: (1) \textit{Off-the-shelf Model Selection} via priority-mapped recommendation matrices, (2) \textit{Architectural Customization} via peak SRAM activation budgeting and resolution-first scaling, and (3) \textit{Runtime Optimization} via algorithm--engine co-design and uniform integer execution.
\end{enumerate}

\subsection{Research Questions and Contributions}

Guided by the 5-level framework, this survey addresses two central research questions:
\begin{itemize}
    \item \textbf{RQ1 (Hardware and Software Governance):} \textit{How do diverse edge hardware tiers (Level 1) and deployment software ecosystems (Level 3) govern deep learning vision performance and operator compatibility across the compute spectrum?}
    \item \textbf{RQ2 (Hardware-Aware Model Selection and Design):} \textit{Given hard physical constraints on SRAM, Flash, and accelerator dataflow (Level~1), how should practitioners select off-the-shelf vision architectures (Level~2) or adapt model topologies and runtime configurations to satisfy specific deployment priority regimes (Level~5)?}
\end{itemize}

The primary contributions of this survey are synthesized as follows:
\begin{enumerate}
    \item \textbf{Unified 5-Level Deployment Framework:} We formulate a comprehensive, multi-layer framework (Fig.~\ref{fig:framework_overview}) that systematically connects hardware constraints, model architectures, runtime compilation pipelines, empirical performance benchmarks, and actionable design decisions.
    \item \textbf{Multi-Tier Hardware and Software Taxonomy:} We establish an in-depth taxonomy covering mid-range accelerators (GPUs, NPUs, TPUs), dedicated on-die MPU/SoCs, and bare-metal MCUs, detailing their computational mechanics, memory walls, and runtime toolchains.
    \item \textbf{Cross-Platform Benchmark Synthesis:} We aggregate and analyze empirical benchmark data from multiple academic and vendor sources across image classification and object detection, evaluating accuracy, latency, and memory footprints across platforms spanning from 25\,W down to milliwatt MCUs.
    \item \textbf{Actionable Decision Matrices and Design Principles:} We deliver concrete model selection matrices mapped by hardware tier and operational priority (Speed, Balanced, Accuracy), accompanied by foundational design principles (peak SRAM activation budgeting, resolution-first detector scaling, and algorithm--engine co-design).
    \item \textbf{Critical Challenges and Future Roadmap:} We identify six critical deployment bottlenecks—including toolchain fragmentation, operator CPU fallbacks, and the peak activation memory wall—and outline future frontiers in on-device continual learning, unified co-design, and sustainable Edge AI.
\end{enumerate}

\subsection{Survey Organization}

The remainder of this survey is structured as follows: Section~\ref{sec:hardware} covers Level~1 -- Hardware Devices and Resource Constraints; Section~\ref{sec:model} analyzes Level~2 -- Vision Model Space and Scaling Knobs; Section~\ref{sec:software} investigates Level~3 -- Deployment Software Ecosystem and Runtimes; Section~\ref{sec:benchmark} synthesizes Level~4 -- Empirical Benchmarks and Performance Analysis; Section~\ref{sec:framework} establishes Level~5 -- Actionable Design Decisions and Model Mapping; Section~\ref{sec:challenge} examines open deployment challenges and future research directions; and Section~\ref{sec:conclusion} concludes the survey.

\section{Hardware Devices}\label{sec:hardware}
To address the core research questions regarding hardware governance and model deployment constraints, this section establishes a structured taxonomy of edge computing platforms. Rather than evaluating individual boards in isolation, we categorize edge hardware into three distinct tiers based on computational capacity, memory hierarchy, and runtime execution model: (i) \textbf{Mid-Range Devices}, (ii) \textbf{Dedicated Embedded MPUs/SoCs}, and (iii) \textbf{Microcontrollers (MCUs)}. For each tier, we analyze representative platforms, architectural dataflow mechanics, and critical hardware parameters that govern vision model execution.

\subsection{Overview of Computing Architecture Classification}

Before detailing specific devices, it is necessary to establish a taxonomy at the computational architecture level, because the fundamental differences between edge devices stem from how data is moved and reused during computation—a direct determinant of energy efficiency (TOPS/W) and software flexibility. Hardware accelerators for deep learning broadly fall into four architectural paradigms: general-purpose CPUs, GPUs, specialized ASICs/NPUs, and reconfigurable FPGAs.

\subsubsection{CPU}

In the classic Von Neumann architecture on which CPUs are based, computation and memory units are physically separated, requiring data to move continuously across the system bus for each arithmetic operation. For convolution workloads—which necessitate extensive reuse of weights and feature activations—data movement constitutes a major bottleneck, frequently accounting for over 60\% of total system energy consumption \cite{sestito2025trim}. This represents an intrinsic architectural limitation of CPUs for deep learning, although CPUs remain the ubiquitous baseline for small MCUs due to simplicity, low manufacturing cost, and general programmability.

\subsubsection{GPU}

Graphics Processing Units (GPUs) overcome CPU throughput bottlenecks via massive fine-grained parallelism—utilizing thousands of lightweight Single Instruction, Multiple Threads (SIMT) cores operating concurrently on data matrices \cite{raja2024systolic}. While GPUs achieve high raw arithmetic throughput (GFLOPS/TFLOPS), they remain bound by the Von Neumann memory wall when accessing shared off-chip DRAM. Nonetheless, GPUs retain the highest software versatility across edge platforms, seamlessly executing dynamic networks, high-resolution feature extractors, and diverse precision formats (FP32, FP16, INT8) via unified SDKs like NVIDIA TensorRT.

\subsubsection{ASIC/NPU}

\begin{figure}[htbp]
    \centering
    \includegraphics[width=\linewidth]{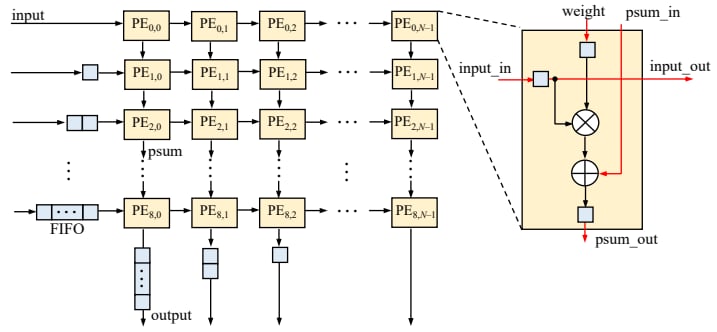}
    \caption{A weight-stationary systolic array architecture, where inputs move left to right and partial sums (psums) move top to bottom across the PE grid. Figure adapted from Sestito et al.~\cite{sestito2025trim}, Fig.~2(a).}
    \label{fig:systolic_array}
\end{figure}

Unlike general-purpose processors, Application-Specific Integrated Circuits (ASICs) and Neural Processing Units (NPUs) dedicated to neural network inference frequently abandon the Von Neumann model in favor of spatial dataflow architectures. In a dataflow design (such as a systolic array or spatial processing element grid), input features and weights stream synchronously through an interconnected array of processing elements, enabling operands to be reused across successive multiply-accumulate (MAC) operations with minimal DRAM accesses \cite{sestito2025trim}, as illustrated in Fig.~\ref{fig:systolic_array}. Edge inference accelerators are commonly classified by their internal compute mechanisms into ALU-based, dataflow-based, and sparsity-aware architectures \cite{capra2020updated}.

\subsubsection{FPGA}

Field-Programmable Gate Arrays (FPGAs) offer reconfigurable logic fabrics that allow hardware designers to synthesize fully custom dataflow pipelines (e.g., weight-stationary, output-stationary, or row-stationary data paths) tailored to specific neural network topologies. While FPGAs achieve exceptional energy efficiency and deterministic low latency \cite{li2025dataflow, venieris2018toolflows}, they demand specialized hardware description language (HDL/HLS) design effort and compiler toolchains compared to standard GPU and ASIC development flows.

The architectural taxonomy outlined above directly maps to the three commercial edge hardware tiers examined throughout this survey: (i) \textbf{Mid-Range Devices}, (ii) \textbf{Dedicated Embedded MPUs/SoCs}, and (iii) \textbf{Microcontrollers (MCUs)}.

\subsection{Mid-Range Devices}\label{sec:midrange_devices}

Mid-range edge platforms encompass embedded single-board computers and system-on-modules (SoMs) equipped with dedicated hardware accelerators (GPUs, NPUs, or dataflow ASICs) running full operating systems (typically Linux). Operating within a 5--25\,W thermal and power envelope, this tier delivers compute capacities spanning from under 1\,TOPS to over 100\,TOPS, enabling real-time execution of medium-to-large vision models.

\begin{figure*}[htbp]
  \centering
  \subfloat[Jetson Nano 4GB\label{fig:jetson_nano}]{\includegraphics[height=2.7cm,keepaspectratio]{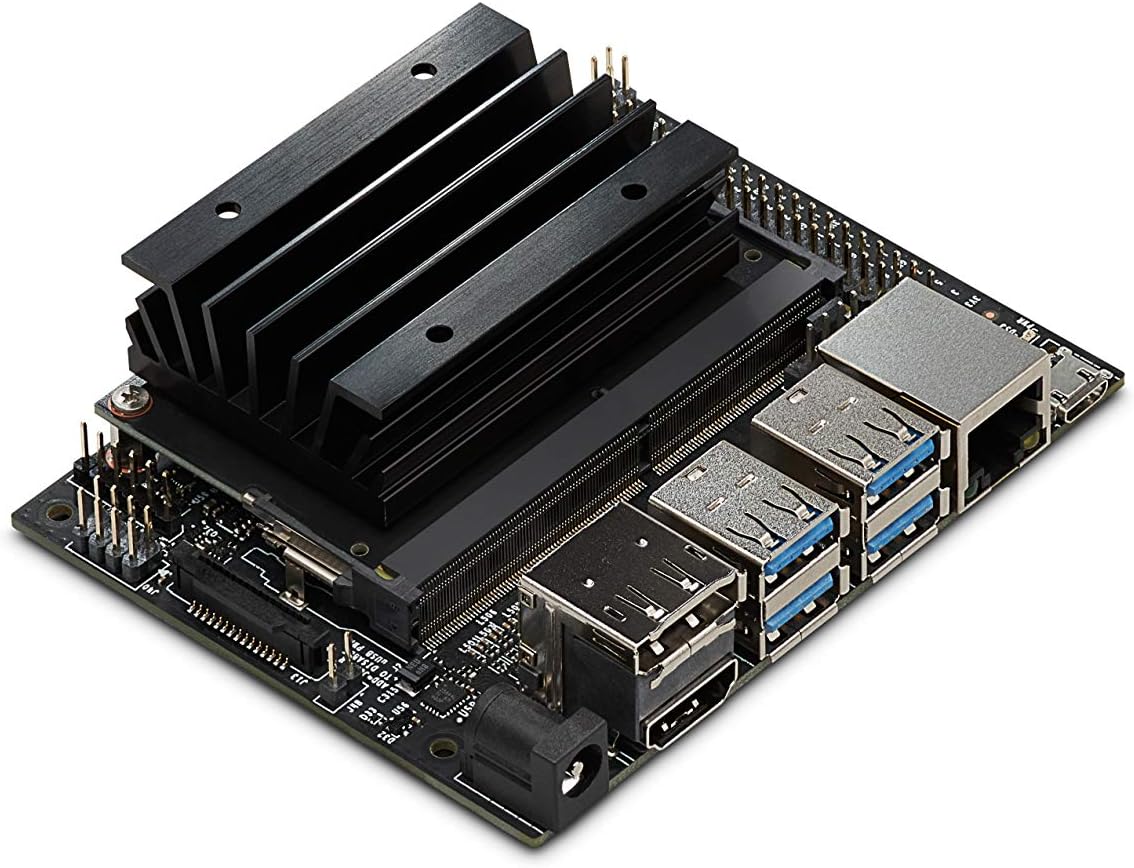}}
  \hfil
  \subfloat[Jetson Orin Nano\label{fig:jetson_orin}]{\includegraphics[height=2.7cm,keepaspectratio]{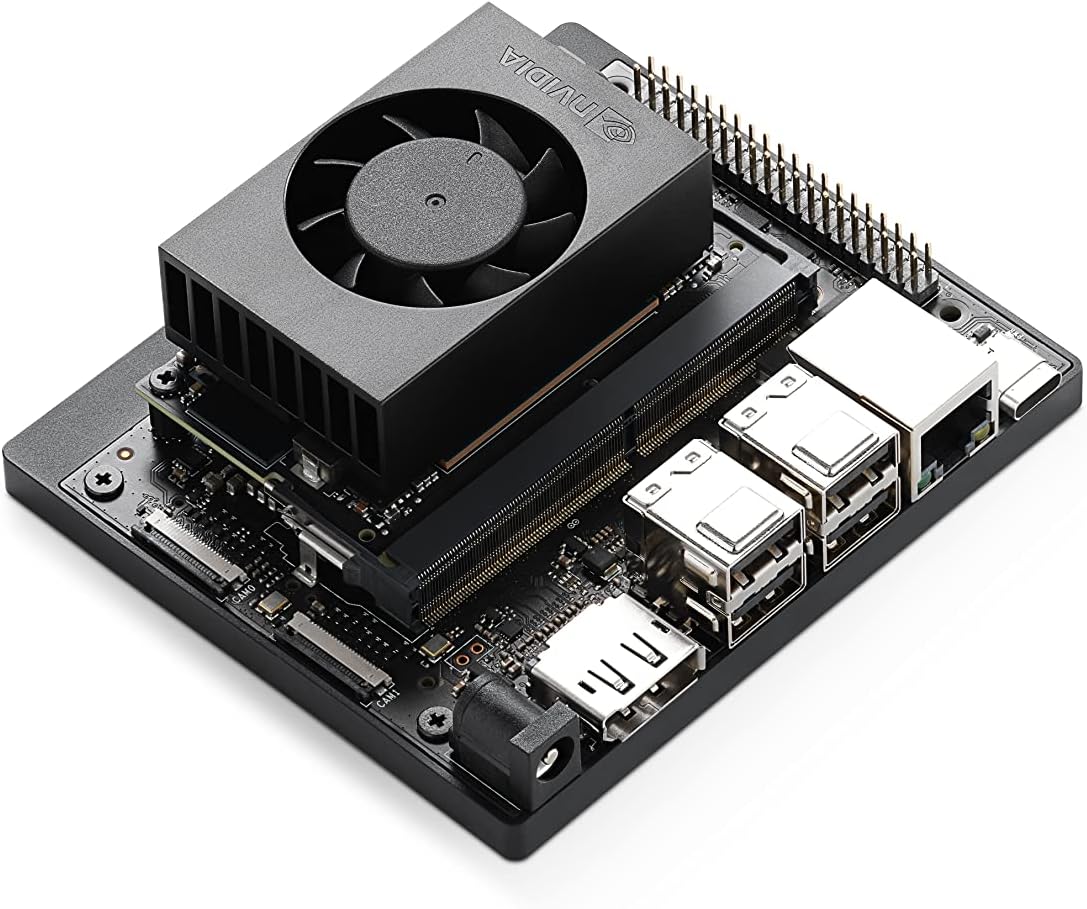}}
  \hfil
  \subfloat[Coral Dev Board\label{fig:coral_dev}]{\includegraphics[height=2.7cm,keepaspectratio]{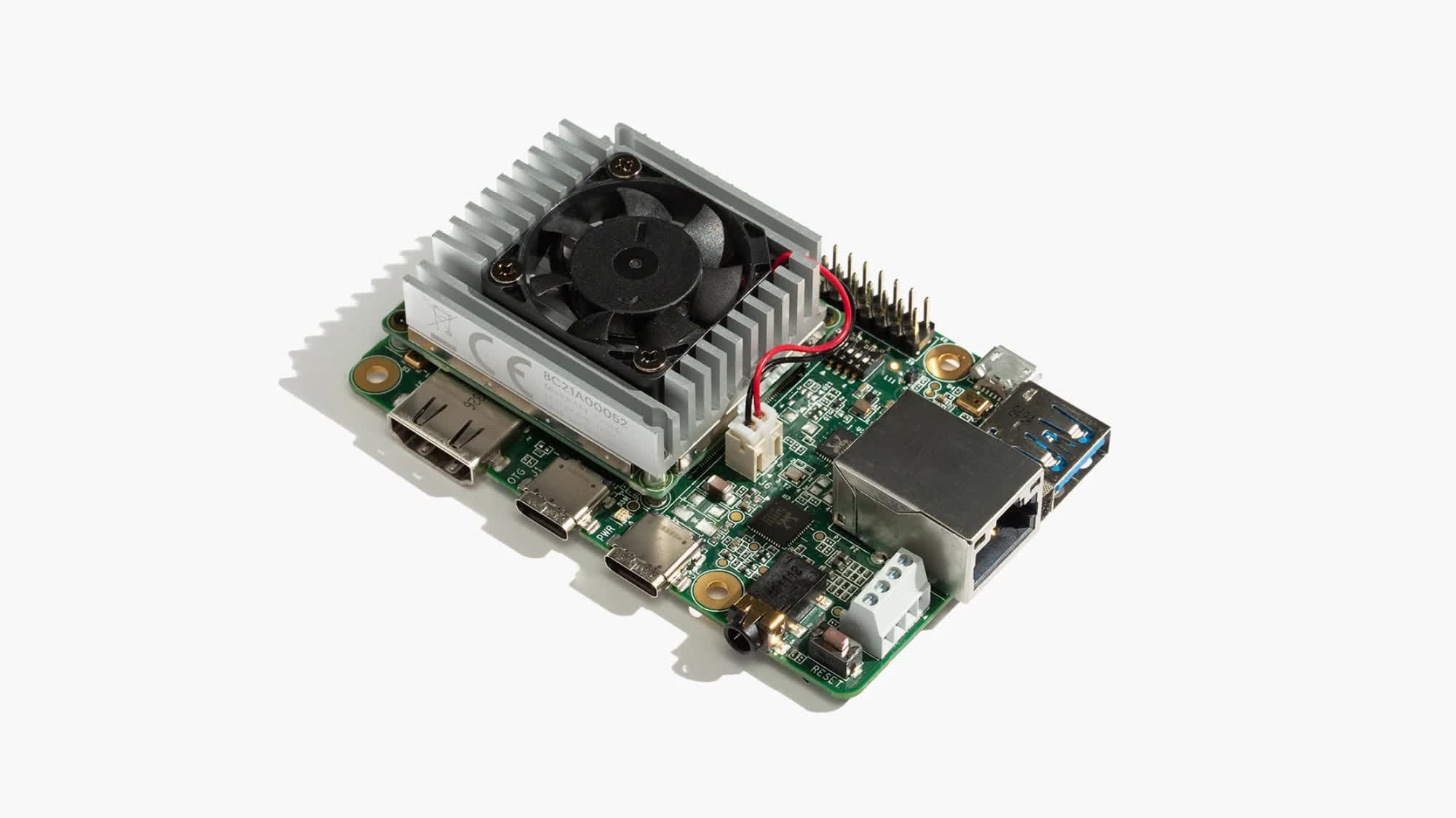}}
  \\[6pt]
  \subfloat[Coral USB Accelerator\label{fig:coral_usb}]{\includegraphics[height=2.7cm,keepaspectratio]{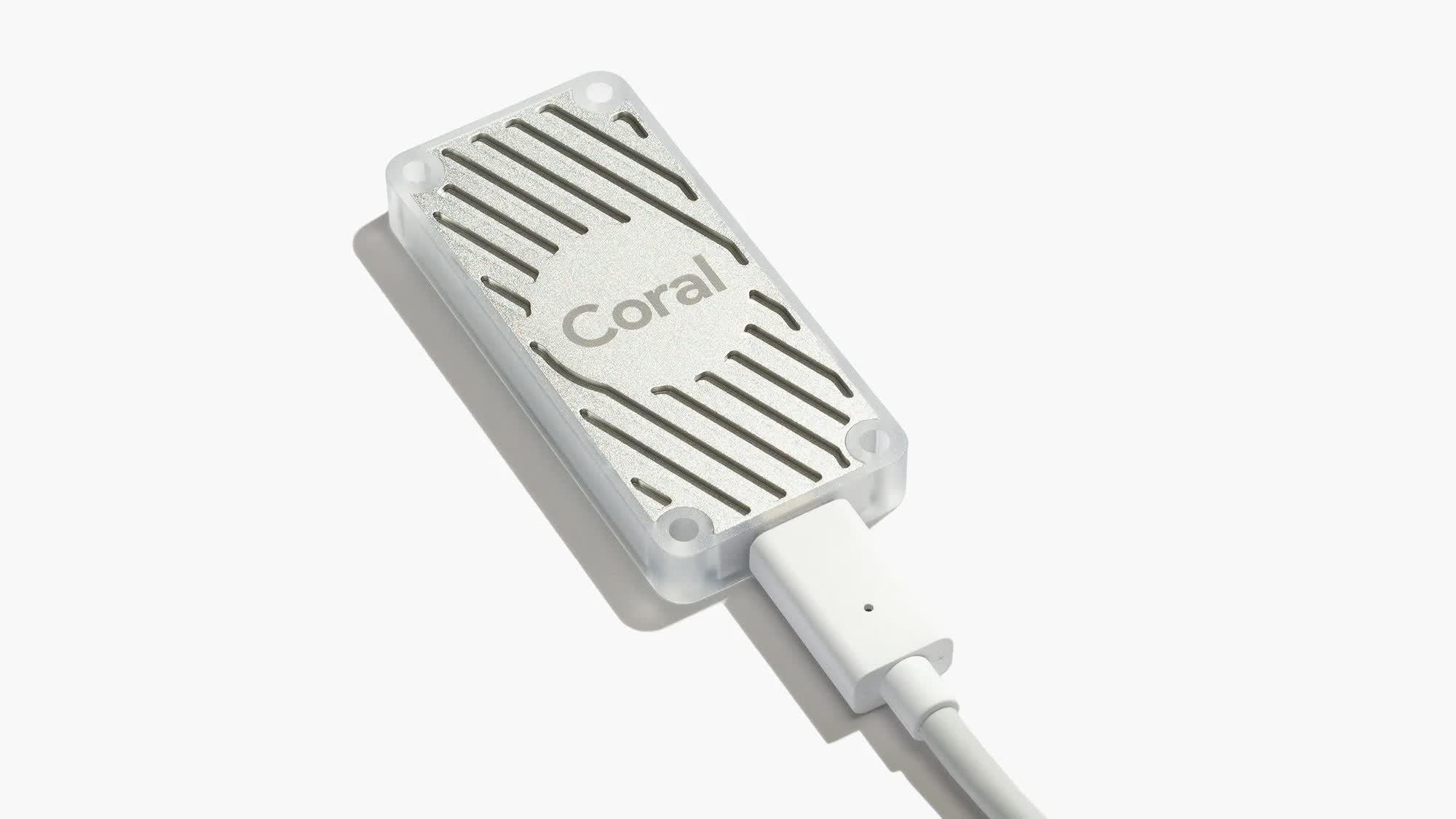}}
  \hfil
  \subfloat[Hailo-8 M.2 Module\label{fig:hailo8_m2}]{\includegraphics[height=2.7cm,keepaspectratio]{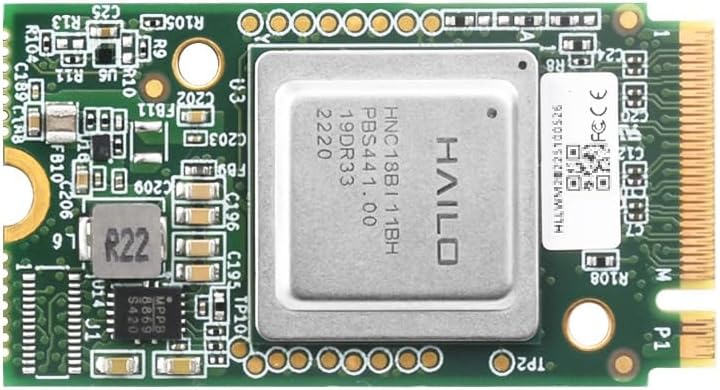}}
  \hfil
  \subfloat[Raspberry Pi 5 (8GB)\label{fig:rpi5}]{\includegraphics[height=2.7cm,keepaspectratio]{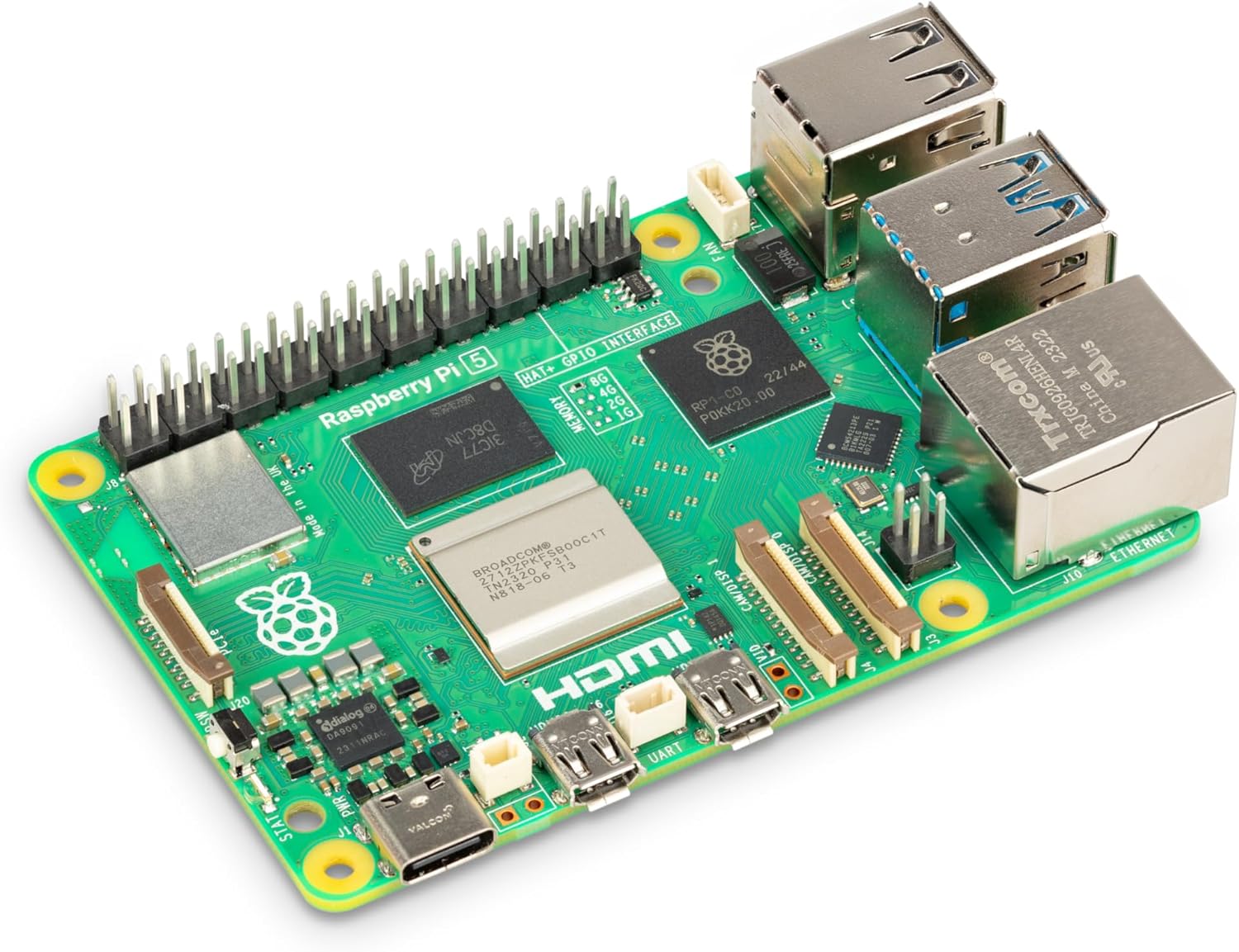}}
  \caption{Representative mid-range Edge AI hardware platforms and accelerator configurations: (a) NVIDIA Jetson Nano 4GB (Maxwell GPU), (b) NVIDIA Jetson Orin Nano (Ampere GPU), (c) Google Coral Dev Board (Edge TPU ASIC), (d) Google Coral USB Accelerator (Edge TPU USB dongle), (e) Hailo-8 M.2 AI acceleration module (26~TOPS dataflow ASIC), and (f) Raspberry Pi 5 8GB single-board computer acting as a host system for discrete accelerators. (Image sources: (a, b, e, f) Amazon product listings; (c, d) Google Coral official product pages).}
  \label{fig:midrange_devices}
\end{figure*}

Architecturally, devices in this tier follow three distinct accelerator integration paradigms, as illustrated in Fig.~\ref{fig:midrange_devices}:
\begin{enumerate}
    \item \textbf{Integrated On-Board GPU Systems}: Platforms such as the NVIDIA Jetson Nano (Fig.~\ref{fig:jetson_nano}) and the Jetson Orin Nano (Fig.~\ref{fig:jetson_orin}) integrate multi-core CPUs with programmable GPU streaming multiprocessors on a shared high-bandwidth memory bus, providing exceptional operator flexibility and high floating-point throughput.
    \item \textbf{Self-Contained Dataflow ASIC Dev Boards}: Standalone development boards such as the Google Coral Dev Board (Fig.~\ref{fig:coral_dev}) integrate an on-board host CPU alongside a custom tensor processing unit (Edge TPU) optimized specifically for quantized INT8 matrix operations.
    \item \textbf{Discrete Accelerators Paired with Host SBCs}: Modular accelerators—such as the Google Coral USB Accelerator dongle (Fig.~\ref{fig:coral_usb}) or the Hailo-8 M.2 AI acceleration module (Fig.~\ref{fig:hailo8_m2})—are interfaced via USB 3.0 or PCIe/M.2 slots to general-purpose host single-board computers, such as the Raspberry Pi 5 (Fig.~\ref{fig:rpi5}).
\end{enumerate}

This architectural dichotomy directly impacts empirical benchmark evaluation \cite{baller2021deepedgebench}: while integrated boards (e.g., Jetson Nano, Coral Dev Board) execute inference within a unified SoC fabric, discrete accelerator throughput is heavily bounded by the host CPU's capability to handle image pre/post-processing, frame orchestration, and interconnect bus transfer overhead (e.g., USB 3.0 latency vs. direct PCIe Gen2/3 lanes). Table~\ref{tab:benchmark_device_midrange} summarizes the representative mid-range platforms evaluated in our benchmarks.

\begin{table}[htbp]
    \centering
    \caption{Mid-range devices evaluated in benchmarks}
    \label{tab:benchmark_device_midrange}
    \resizebox{0.48\textwidth}{!}{
    \begin{tabular}{lllc}
        \toprule
        \textbf{Device} & \textbf{CPU} & \textbf{AI Accelerator} & \textbf{AI Peak Perf.} \\
        \midrule
        Jetson Nano~\cite{Tobiasz2023EdgeDI,Alqahtani2024ACE} & 4-core Cortex-A57 MPCore      & 128-core Maxwell & 472 GFLOPS~\cite{nvidia_jetson_nano_specs} \\
        Jetson Orin Nano~\cite{Alqahtani2024ACE}              & 6-core Cortex-A78AE           & 512-core Ampere  & 34 TOPS~\cite{nvidia_jetson_orin_specs} \\
        Coral Dev Board~\cite{coral_edgetpu_benchmarks}       & 4-core Cortex-A53 @ 1.5GHz    & Edge TPU         & 4 TOPS~\cite{google_coral_dev_board_datasheet} \\
        Coral USB~\cite{coral_edgetpu_benchmarks}             & Intel Xeon Gold 6154 @ 3.0GHz & Edge TPU         & 4 TOPS~\cite{google_coral_usb_accelerator_datasheet} \\
        Hailo Accelerator~\cite{hailo_model_explorer}         & Intel Core i5-9400 @ 2.9GHz   & Hailo-8          & 26 TOPS~\cite{hailo_8_ai_accelerator} \\
        RPi5 + Coral USB~\cite{Alqahtani2024ACE}              & 4-core Cortex-A76 @ 2.4GHz    & Edge TPU         & 4 TOPS~\cite{google_coral_usb_accelerator_datasheet} \\
        RPi5 + AI HAT+~\cite{Alqahtani2024ACE}                & 4-core Cortex-A76 @ 2.4GHz    & Hailo-8          & 26 TOPS~\cite{hailo_8_ai_accelerator} \\
        \bottomrule
    \end{tabular}
    }
\end{table}

In terms of technology generation, the mid-range performance spectrum reflects two evolutionary waves:
1) \textit{First-generation edge accelerators} (e.g., Maxwell-based GPU on Jetson Nano in Fig.~\ref{fig:jetson_nano}, first-generation Edge TPU on Coral Dev Board in Fig.~\ref{fig:coral_dev} and Coral USB in Fig.~\ref{fig:coral_usb}), delivering $\le 4$\,TOPS / 472\,GFLOPS with restricted INT8 operator subsets or limited memory bandwidth.
2) \textit{Modern edge accelerators} (e.g., Ampere architecture on Jetson Orin Nano in Fig.~\ref{fig:jetson_orin}, and high-density spatial dataflow architectures on the Hailo-8 M.2 module in Fig.~\ref{fig:hailo8_m2}), delivering 20--40+\,TOPS.

This generational gap is not simply linear in power consumption, but stems from architectural enhancements and advanced compiler toolchains (TensorRT with FP16/INT8 kernel fusion, Hailo Dataflow Compiler). Crucially, this defines the feasibility boundary for Vision Transformers (ViTs) and modern hybrid backbones: while modern devices possess sufficient memory bandwidth and operator support for attention mechanisms, older architectures (e.g., Jetson Nano, Coral TPU) encounter severe memory bottlenecks or unsupported operator fallbacks.

\subsection{Dedicated Embedded MPU/SoC}\label{sec:MPU/SoC}

To overcome the performance and power trade-offs of standalone MCUs and mid-range accelerators, an emerging class of dedicated embedded MPUs and SoCs integrates a neural accelerator block directly on the same die as the CPU. This architectural integration miniaturizes the spatial dataflow principles of discrete ASICs down to the milliwatt power envelope. The dedicated NPU block handles computationally intensive, highly repetitive 2D convolutions with maximal on-chip data reuse, while the primary CPU core (or a compact RISC-V cluster) manages system orchestration, sensor I/O, and non-linear pre/post-processing operations.

Unlike discrete accelerator configurations that require dedicated PCB packages, external DRAM, and high-power PCIe/USB PHY interfaces, on-die NPU integration communicates via wide internal SoC interconnects. This architectural approach delivers two simultaneous advantages: (i) eliminating the power and latency overhead of peripheral communication and OS daemon management—thereby sustaining milliwatt-scale operation suitable for battery-powered systems, and (ii) preserving the spatial data reuse efficiency of dedicated dataflow hardware—yielding significantly higher throughput than sequential CPU execution.

Prominent commercial examples of this architectural convergence include STMicroelectronics' STM32N6, which couples an ARM Cortex-M55 core with the proprietary Neural-ART NPU \cite{carboni2026tiny}; GreenWaves Technologies' GAP9, featuring a multi-core RISC-V compute cluster paired with the NE16 neural accelerator delivering tens of mW efficiency \cite{muller2024gap9shield}; and the Himax WiseEye2 (HX6538), which pairs an ARM Cortex-M55 with Arm's Ethos-U55 microNPU for always-on vision sensing \cite{millar2025benchmarking}.

\subsection{Microcontrollers (MCUs)}\label{sec:MCUs}

Unlike the mid-range and dedicated MPU/SoC tiers, the microcontroller (MCU) tier consists of ultra-low-power devices built entirely around general-purpose CPU cores—primarily ARM Cortex-M series (Cortex-M0+/M3/M4/M7/M33) or Espressif ESP32 families—operating without dedicated neural acceleration hardware. Running bare-metal firmware or lightweight real-time operating systems (RTOS), MCUs operate within strict milliwatt power budgets and severely constrained memory footprints (tens to hundreds of kilobytes of SRAM and a few megabytes of internal Flash). This tier constitutes the core focus of TinyML and TinyDL deployment \cite{Somvanshi2025FromTM, lin2023tiny}.

\begin{figure}[htbp]
    \centering
    \includegraphics[width=0.6\linewidth]{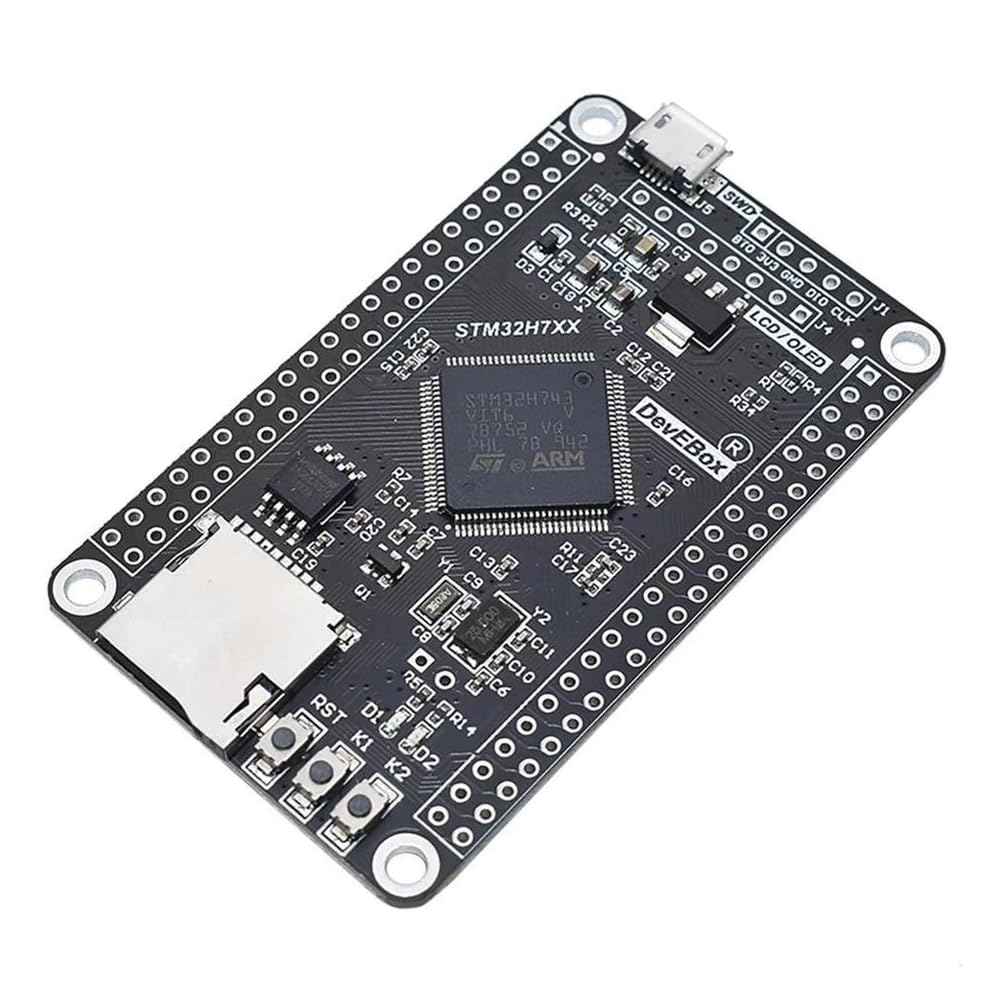}
    \caption{STM32H743 development board (ARM Cortex-M7 @ 480\,MHz, 512\,KB internal SRAM, 2\,MB Flash), representing a high-performance general-purpose MCU platform for TinyML vision inference. (Image source: Amazon product listing).}
    \label{fig:stm32h743}
\end{figure}

An exemplary high-performance microcontroller platform is the STM32H743 development board shown in Fig.~\ref{fig:stm32h743}, powered by an ARM Cortex-M7 core running at 480\,MHz with 512\,KB internal SRAM and 2\,MB Flash. Because MCUs lack hardware matrix multipliers, neural network convolutions are executed sequentially on the CPU core using optimized software kernels, such as TensorFlow Lite for Microcontrollers (TFLM) or CMSIS-NN \cite{lai2018cmsis}, leveraging the Armv7E-M / Armv8-M DSP and SIMD instruction sets. Deep quantization (typically INT8) and extreme parameter pruning are essential prerequisites for fitting models into MCU SRAM budgets. Recent benchmarking indicates that general-purpose MCUs without dedicated hardware acceleration (e.g., STM32H7A3ZI, ESP32-S3) exhibit significantly lower energy efficiency during vision workloads compared to NPU-augmented platforms \cite{millar2025benchmarking}.

Table~\ref{tab:bench_device_mcu} details the MCU platforms evaluated in our benchmarks, categorized across CPU core capabilities and internal/external memory limits. In terms of processing cores, devices featuring Cortex-M4 (e.g., STM32F469, STM32L4R9I-DISCO) operate at lower clock frequencies than Cortex-M7 devices (e.g., STM32H743 in Fig.~\ref{fig:stm32h743}, STM32F746, NUCLEO-H743ZI, STM32F769I-DISCO) which benefit from dual-issue superscalar pipelines. Heterogeneous platforms such as the STM32H747I-DISCO combine a Cortex-M7 and a Cortex-M4 core on the same die to segregate real-time control from inference workloads. Both Cortex-M4 and Cortex-M7 cores support the ARM DSP/SIMD instruction set, which CMSIS-NN utilizes to accelerate quantized convolution by up to $4.6\times$ in throughput and $4.9\times$ in energy efficiency over unoptimized C implementations \cite{lai2018cmsis}.

\begin{table}[htbp]
    \centering
    \caption{Microcontrollers evaluated in benchmarks}
    \label{tab:bench_device_mcu}
    \resizebox{0.48\textwidth}{!}{
    \begin{tabular}{llrrrr}
        \toprule
        \multirow{2}{*}{\textbf{Device}} & \multirow{2}{*}{\textbf{ARM Cortex Core}} & \multicolumn{2}{c}{\textbf{Internal Mem (KB)}} & \multicolumn{2}{c}{\textbf{External Mem (KB)}} \\
        \cmidrule{3-4} \cmidrule{5-6}
        & & \textbf{SRAM} & \textbf{Flash} & \textbf{SDRAM} & \textbf{ExtFlash} \\
        \midrule
        STM32H743~\cite{lin2023tiny}              & M7 @ 480MHz    & 512  & 2048 & -     & - \\
        STM32F469~\cite{Sadiq2022TinyOpsIS}       & M4 @ 180MHz    & 256  & 1024 & 8192  & 8192 \\
        STM32F746~\cite{Sadiq2022TinyOpsIS}       & M7 @ 216MHz    & 320  & 1024 & 8192  & 8192 \\
        NUCLEO-H743ZI~\cite{Sah2024MCUBenchAB}    & M7 @ 480MHz    & 1024 & 2048 & -     & - \\
        B-U585I-IOT02A~\cite{Sah2024MCUBenchAB}   & M33 @ 160MHz   & 768  & 2048 & -     & 65536 \\
        STM32F469I-DISCO~\cite{Sah2024MCUBenchAB} & M4 @ 180MHz    & 384  & 2048 & 16384 & 16384 \\
        STM32F769I-DISCO~\cite{Sah2024MCUBenchAB} & M7 @ 216MHz    & 512  & 2048 & 16384 & 65536 \\
        STM32H573I-DK~\cite{Sah2024MCUBenchAB}    & M33 @ 250MHz   & 640  & 2048 & -     & 65536 \\
        STM32H747I-DISCO~\cite{Sah2024MCUBenchAB} & M4+M7 @ 400MHz & 704  & 1024 & 8192  & 131072 \\
        STM32L4R9I-DISCO~\cite{Sah2024MCUBenchAB} & M4 @ 120MHz    & 640  & 2048 & -     & 65536 \\
        \bottomrule
    \end{tabular}
    }
\end{table}

In terms of memory hierarchy, on-chip SRAM is universally the most critical bottleneck on MCUs because it must accommodate the peak intermediate activation buffers during inference. While internal Flash and external memories (e.g., 128\,MB external Flash on STM32H747I-DISCO) alleviate model weight storage constraints, external memory accesses incur severe latency and energy penalties, and external RAM cannot substitute for low-latency on-chip SRAM. This hard activation boundary necessitates specialized memory-aware neural architectures and runtime tiling frameworks (Section~\ref{sec:framework}).

\subsection{Summary of Hardware Constraints}

\begin{table*}[htbp]
\centering
\small
\setlength{\tabcolsep}{5pt}
\renewcommand{\arraystretch}{1.5}
\caption{Comparison of memory and storage resources between AI platforms and neural network models (Source: Lin et al.~\cite{lin2020mcunet})}
\label{tab:ai_hardware_comparison}
\begin{tabular}{l c c c c c c c c c}
\toprule
 & \textbf{\begin{tabular}[c]{@{}c@{}}Cloud AI\\ (NVIDIA V100)\end{tabular}} & $\xrightarrow{\text{\color{purple}4}\times}$ & \textbf{\begin{tabular}[c]{@{}c@{}}Mobile AI\\ (iPhone 11)\end{tabular}} & $\xrightarrow{\text{\color{purple}3100}\times}$ & \textbf{\begin{tabular}[c]{@{}c@{}}Tiny AI\\ (STM32F746)\end{tabular}} & & \textbf{ResNet-50} & \textbf{MobileNetV2} & \textbf{\begin{tabular}[c]{@{}c@{}}MobileNetV2\\ (int8)\end{tabular}} \\
\midrule
\textbf{Memory} & 16 GB & & 4 GB & & \textbf{320 kB} & {\color{purple}$\leftarrow$gap$\rightarrow$} & 7.2 MB & 6.8 MB & 1.7 MB \\
\textbf{Storage} & TB$\sim$PB & $\xrightarrow{\text{\color{purple}1000}\times}$ & $>$64 GB & $\xrightarrow{\text{\color{purple}64000}\times}$ & \textbf{1 MB} & {\color{purple}$\leftarrow$gap$\rightarrow$} & 102MB & 13.6 MB & 3.4 MB \\
\bottomrule
\end{tabular}
\end{table*}

Synthesizing the multi-tier analysis reveals distinct divergence across computational, memory, and energy axes:

\textbf{Computational Architecture}: Mid-range devices rely on SIMT GPUs or spatial dataflow ASICs running a full Linux OS, with discrete accelerators communicating over PCIe or USB 3.0 buses (Fig.~\ref{fig:midrange_devices}). Dedicated embedded MPUs/SoCs bridge this gap by integrating NPU dataflow engines with CPU cores on a single die via low-latency internal buses. Microcontrollers represent the simplest architectural tier, relying entirely on sequential CPU execution without hardware matrix acceleration.

\textbf{Memory Hierarchy and the Capacity Gap}: As quantified in Table~\ref{tab:ai_hardware_comparison}, memory and storage resources contract non-linearly across tiers \cite{lin2020mcunet}. Transitioning from cloud accelerators (NVIDIA V100, 16\,GB) to mobile devices shrinks memory by $4\times$, but transitioning from mobile to a standard microcontroller (e.g., STM32F746 with 320\,KB SRAM) reduces memory by over $3100\times$ and storage by $64000\times$. This extreme disparity illustrates why standard mobile models cannot be scaled down to MCUs by conventional post-training quantization alone: even an INT8-quantized MobileNetV2 requires 1.7\,MB of peak working SRAM—far exceeding the 320\,KB SRAM capacity of typical microcontrollers.

\textbf{Power and Thermal Budgets}: Mid-range platforms operate at 5--25\,W (Table~\ref{tab:benchmark_device_midrange}), requiring active cooling or sizable passive heatsinks and continuous power supplies. Dedicated MPUs/SoCs operate in the tens of milliwatts, enabling untethered, battery-operated deployment. Standard MCUs maintain low milliwatt-level consumption due to low clock frequencies (120--480\,MHz) and lack of high-power hardware acceleration logic.

These three axes interact tightly: a model architecture that satisfies computational throughput requirements may still fail if its activation memory exceeds SRAM limits (Table~\ref{tab:ai_hardware_comparison}), or if sustained thermal dissipation limits are breached. Consequently, hardware-aware model selection and design must evaluate seven essential parameters:
\begin{enumerate}
    \item \textbf{Accelerator Architecture}: GPU SIMT, spatial dataflow ASIC, or general-purpose CPU.
    \item \textbf{Computational Capacity}: Peak GFLOPS/TOPS (Table~\ref{tab:benchmark_device_midrange}) coupled with toolchain kernel fusion efficiency.
    \item \textbf{Internal SRAM Budget}: Peak activation buffer capacity (Table~\ref{tab:bench_device_mcu}), the primary constraint on MPUs and MCUs.
    \item \textbf{Flash/Storage Capacity}: Internal and external non-volatile weight storage limits (Table~\ref{tab:bench_device_mcu}).
    \item \textbf{Power Envelope}: Watt-scale for mid-range systems versus milliwatt-scale for MCUs/MPUs.
    \item \textbf{Input Resolution}: Governs both inference latency and peak intermediate activation memory.
    \item \textbf{Software Ecosystem}: Runtime compiler support, operator coverage, and quantization compatibility (Section~\ref{sec:software}).
\end{enumerate}
This multi-dimensional parameter space forms the foundation for the unified hardware-to-model mapping framework presented in Section~\ref{sec:framework}.

\section{Edge Vision Models}\label{sec:model}
This section presents some commonly benchmarked CNN architectures, commonly used optimization techniques, and an outline of recent research on Vision Transformer architecture at the edge.

\subsection{CNN architectures}

MobileNet is a family of lightweight architectures for mobile devices and embedded vision applications, developed by Google. MobileNetV1~\cite{Howard2017MobileNetsEC} uses depthwise separable convolution to reduce parameters and computation while maintaining accuracy, introducing the width multiplier $\alpha$ and resolution multiplier $\rho$ to tune latency-accuracy trade-offs. MobileNetV2~\cite{Sandler2018MobileNetV2IR} introduces inverted residuals along with linear bottlenecks, which improve representational capacity and memory efficiency. MobileNetV3~\cite{Howard2019SearchingFM} advances this series by combining hardware-aware neural architecture search (NAS) with lightweight Squeeze-and-Excitation attention and hardware-friendly hard-swish activation functions, providing optimized Large and Small variants for power-constrained platforms.

Inception is also an architecture family developed by Google. InceptionV1~\cite{Szegedy2014GoingDW}, also known as GoogLeNet, significantly increases network depth and breadth without increasing computational cost through the efficient use of 1$\times$1 convolutions. Besides, quality is also optimized thanks to multi-scale processing. InceptionV3~\cite{Szegedy2015RethinkingTI} continues to improve network scaling efficiency by decomposing convolutional layers and applying regularization techniques to optimize the training process.

ResNet~\cite{He2015DeepRL} is a widely used architecture in computer vision tasks. With deep residual learning and shortcut connections, it becomes easier to train very deep models, thereby achieving higher accuracy.

MnasNet~\cite{tan2019mnasnet} is an architecture created using neural automatic architecture search (NAS) for mobile devices with the goal of balancing accuracy and latency. Similarly, ProxylessNAS~\cite{cai2018proxylessnas} also uses NAS but optimizes the found architecture according to the task and target hardware, while reducing the search cost many times compared to previous methods.

MCUNet~\cite{lin2020mcunet} is an architecture specifically designed for microcontrollers with extremely limited resources. Therefore, to maximize performance on these devices, MCUNet combines automated architecture search (TinyNAS) with a memory-optimized inference library (TinyEngine). From there, the search space can be expanded to produce larger architectures.

\subsection{Object detectors}

YOLO (You Only Look Once) is a family of famous single-stage object detectors with fast inference speed, suitable for real-time applications. YOLOv3~\cite{Redmon2018YOLOv3AI} uses the Darknet-53 architecture and multi-scale prediction. After that, the development line divided into many branches. On the academic side, YOLOv6~\cite{Li2022YOLOv6AS, Li2023YOLOv6VA} is developed by Meituan, with improvements in architecture (EfficientRep - re-parameterization from RepVGG) and training strategy; this version offers a balance between speed and accuracy. YOLOv7~\cite{Wang2022YOLOv7TB} introduces the E-ELAN architecture, which achieves significant superiority in both speed and accuracy over many previous real-time detectors. In parallel, Ultralytics develops branches YOLOv5~\cite{Ultralytics2020YOLOv5}, YOLOv8~\cite{Ultralytics2023YOLOv8}, YOLO11~\cite{Ultralytics2024YOLO11} and YOLO26~\cite{Ultralytics2026YOLO26}, focusing on ease of use and deployment support cross-platform.

SSD (Single Shot MultiBox Detector)~\cite{Liu2015SSDSS} is another one-stage object detector that uses feature maps at multiple scales to predict bounding boxes and confidence scores of objects with different sizes. However, the VGG16 backbone in~\cite{Liu2015SSDSS} is not effective for inference at the edge, so it is replaced by MobileNet~\cite{Howard2017MobileNetsEC, Sandler2018MobileNetV2IR} and is called SSD-MobileNet~\cite{tensorflow_detection_api}. Another variation of~\cite{Liu2015SSDSS} is SSDLite-MobileDet~\cite{tensorflow_detection_api}, which is a combination of SSDLite~\cite{Sandler2018MobileNetV2IR} and MobileDet~\cite{Xiong2020MobileDetsSF} to optimize mobile device performance.

EfficientDet~\cite{Tan2019EfficientDetSA} is an object detector that uses BiFPN (Bi-directional Feature Pyramid Network) for multi-scale feature fusion, combined with compound scaling to effectively scale the model. EfficientDet-Lite~\cite{google_automl_efficientdet} is a version optimized by Google for mobile and IoT devices, using EfficientNet-Lite~\cite{tensorflow_tpu_efficientnet_lite} as the backbone.

\subsection{Optimization techniques}

While academic research on optimization techniques for Edge AI often refers to model pruning, quantization, knowledge distillation, or tensor decomposition~\cite{lin2023tiny}, in practice, developers often favor simple and easy-to-implement techniques when hardware constraints are not too strict. Specifically, for mid-range hardware devices, since current configurations can run most small and medium models, the commonly used technique is quantization to match the requirements of each hardware~\cite{lin2020mcunet, Sadiq2022TinyOpsIS,Somvanshi2025FromTM}; for example, FP16 for GPUs like NVIDIA Jetson, INT8 for NPUs like Google Coral~\cite{Tobiasz2023EdgeDI}. On the contrary, on devices with extremely limited resources such as microcontrollers, specialized, deep optimization techniques will be applied. MCUNet~\cite{lin2020mcunet} is a typical work for this approach. By co-designing algorithms and systems, MCUNet optimizes the architecture while ensuring that hardware constraints are met. Besides, mobile-optimized architectures, such as MobileNet~\cite{Howard2017MobileNetsEC, Sandler2018MobileNetV2IR}, MnasNet~\cite{tan2019mnasnet}, ProxylessNAS~\cite{cai2018proxylessnas}, can also run on microcontrollers by compressing model size and reducing memory consumption during inference with technique to reduce model width and input image size (width/resolution scaling)~\cite{Howard2017MobileNetsEC, Sandler2018MobileNetV2IR,tan2019mnasnet,cai2018proxylessnas,lin2023tiny, Sadiq2022TinyOpsIS, Sah2024MCUBenchAB}.

\subsection{Vision Transformers}

Vision Transformer (ViT) is a computer vision architecture based on the self-attention mechanism, different from CNN families that extract features according to a local hierarchical structure. Instead of scanning a fixed convolution window across the image, ViT divides the input image into a sequence of patches, linearizes each patch into an embedding, and processes the entire sequence with self-attention layers that allow each patch to interact directly with every other patch from the first layer. This mechanism gives ViT a global view from the beginning, without the limitation of increasing network depth like CNN, and thus achieves superior accuracy compared to CNN architectures of the same parameter scale on many classic computer vision benchmarks such as image classification, object detection, and image segmentation~\cite{Saha2025VisionTO}. However, this same global self-attention mechanism is also the source of computational and memory costs that increase quadratically with the number of input patches, causing native ViT to have significantly higher inference latency and parameter capacity than the lightweight CNN families examined, posing a major challenge when deployed on edge devices with limited computational and memory resources~\cite{papa2024survey}.

Current research on efficient ViT focuses on three groups of techniques: (i) lightweight/hybrid CNN-attention architecture to reduce global self-attention overhead, (ii) model compression (pruning, knowledge distillation, quantization) adapted to the token structure and attention map specific to ViT, and (iii) specialized hardware acceleration for attention matrix multiplication. However, these techniques are currently largely still at the level of theoretical proposals or evaluations on data center GPUs; only a few works mention accelerators specifically for ViT, and almost no works directly verify on edge device layers \cite{Saha2025VisionTO,papa2024survey}.

From an experimental perspective, the effectiveness of the Vision Transformer architecture has recently been verified on smartphones. EfficientFormer repositions ViT's speed-choking operations through direct on-device latency analysis, and results show that the lightest version of this architecture achieves a latency of about 1.6ms on the iPhone 12 when compiled with CoreML, equivalent to the speed of MobilenetV2 but maintaining higher accuracy \cite{liu2023efficientvit}. FastViT re-parameterizes the structure to eliminate memory-wasting skip-connections at inference time, and was benchmarked on the iPhone 12 Pro series \cite{vasu2023fastvit}. MobileViT approaches combine MobileNetV2 convolution blocks and self-attention blocks, targeting mobile devices \cite{mehta2021mobilevit}.

\section{Deployment Software Ecosystem}\label{sec:software}
This section examines the software ecosystem corresponding to the three device groups classified and analyzes cross-platform compatibility challenges.

\subsection{Deployment software ecosystem for mid-range devices}

With the mid-range device group, due to running a full operating system, the software ecosystem is closest to the standard deep learning tool among the three groups, but still requires a separate compilation/optimization step for each hardware architecture. For the NVIDIA Jetson family, TensorRT is the main inference runtime, performing optimizations, precision tuning, and selecting the optimal kernel according to the target GPU architecture. A study directly comparing multiple inference frameworks on Jetson AGX Orin — including PyTorch, ONNX Runtime, TensorRT, Apache TVM, and JAX — shows that TensorRT achieves the lowest latency, while ONNX Runtime, despite having the advantage of portability thanks to the same intermediate format ONNX, on the other hand, is significantly slower with large models \cite{ratul2025accelerating}. For Google Coral, the model must be trained or converted to the quantized TensorFlow Lite format, then compiled separately using Edge TPU Compiler, which is also one of the frameworks directly benchmarked in the cross-device benchmark study \cite{Tobiasz2023EdgeDI, baller2021deepedgebench}. For Hailo, HailoRT serves as the execution runtime, while Hailo Dataflow Compiler takes care of mapping the model's computational graph to the physical dataflow structure of the chip — a compilation process that is significantly more complex than TensorRT or Edge TPU Compiler because it must decide how to allocate each network layer to fixed processing units on the chip.

In addition to the three vendor-specific runtimes mentioned above, Intel's OpenVINO, although not the main tool for the three hardware representatives of this survey, often appears in framework comparison studies as a baseline for Intel hardware (CPU, VPU), and together with TensorRT, ONNX Runtime forms the trio of frameworks most commonly compared in the edge inference benchmark document \cite{lai2018cmsis}. What the entire mid-range software ecosystem has in common is that each runtime achieves its best performance on the hardware for which it was designed but at the expense of the model having to go through a separate conversion, quantization, and compilation step for each deployment target - there is no intermediate format that applies equally effectively across all three platforms, Jetson, Coral, and Hailo.

\subsection{Deployment software ecosystem for low-end devices}

In the dedicated embedded MPU/SoC and microcontroller groups, the absence of a full operating system eliminates the possibility of using the above runtimes, while giving rise to a much more fragmented framework ecosystem. TensorFlow Lite for Microcontrollers (TFLM) is the most commonly used reference runtime, with an OS-agnostic design, no dynamic memory allocation, and static memory that can be configured according to the device's RAM budget \cite{david2021tensorflow}. CMSIS-NN, a kernel library developed by ARM, is often used as a convolution acceleration backend underneath TFLM on Cortex-M cores, directly exploiting this architecture's SIMD/DSP instruction set to accelerate multiplication-addition operations that are the main bottleneck in the absence of hardware accelerators \cite{lai2018cmsis}. At the same time, microcontroller manufacturers also provide their own proprietary toolchains — typically STMicroelectronics' X-Cube-AI (STM32Cube.AI) — that allow direct conversion of the Keras/TensorFlow model to optimized C code for the STM32 family. A study implementing the MicroAI tool directly comparing TFLM, STM32Cube.AI, and a separate optimization framework shows that no framework has absolute superiority on all criteria. criteria (latency, flash capacity, RAM usage), the appropriate choice depends on which criteria are prioritized in the specific implementation problem \cite{novac2021quantization}.

Besides the commercial/industrial frameworks mentioned above, the academic literature also develops specialized runtimes to more fully exploit the limited memory budget of microcontrollers. TinyEngine, designed concurrently with the MCUNet architecture in Section~\ref{sec:model}, is a typical example. Instead of a general interpreter runtime like TFLM, TinyEngine generates model-specific inference code at compile time, eliminating all the overhead of a general interpreter \cite{lin2020mcunet}. The benchmark data in Table~\ref{tab:benchmark_classification_mcu} of this survey clearly illustrates the performance gap between the two approaches: the same MCUNet architecture running on the same STM32H743 device, the version using TinyEngine achieves latency and active memory capacity many times lower than the version using TFLM, even some larger model configurations cause overflow of memory (OOM) when running on TFLM but still execute is on TinyEngine. In a similar vein, TinyOps is another academic runtime, specifically designed to bring ImageNet-scale accuracy models to run on microcontrollers with extremely tight memory budgets, and is also a direct source of benchmark data for STM32F469/STM32F746 devices in Table~\ref{tab:benchmark_classification_mcu} \cite{Sadiq2022TinyOpsIS}.

For the dedicated embedded MPU/SoC group in Section~\ref{sec:MPU/SoC}, the three representatives, STM32N6, GAP9, and Himax WiseEye2, each come with a compilation toolchain for their own integrated NPU — none of the above-mentioned three MCU frameworks (TFLM, CMSIS-NN, X-Cube-AI) directly support model mapping onto these specialized NPUs, making model development for this device group currently still tied to the exclusive ecosystem of each company.

\section{Benchmarks and Performance Analysis}\label{sec:benchmark}
This section synthesizes benchmark data for image classification and object detection tasks on mid-range devices and MCUs from several academic works and official websites from hardware vendors. It analyzes empirical performance trends, hardware bottlenecks, and execution characteristics across hardware classes, establishing the empirical foundation for the hardware-to-model mapping framework presented in Section~\ref{sec:framework}.

Due to differences in experimental settings of sources and implementation processes on each device line, and models are often quantized, benchmark results on different devices will be slightly affected, but can still be used for reference and comparison of models with each other.

\subsection{Image classification}

Benchmark data for image classification task on mid-range devices in Table~\ref{tab:benchmark_classification_midrange} is compiled from Tobiasz et al.~\cite{Tobiasz2023EdgeDI}, Edge TPU performance benchmarks~\cite{coral_edgetpu_benchmarks}, Coral Models for image classification~\cite{coral_image_classification}, and Hailo AI Model Explorer for computer vision~\cite{hailo_model_explorer}. The models benchmarked on Jetson Nano do not include a classification head, so there is no data on acc@1. In addition, the data source on the Coral Dev Board also has no information about accuracy. The remaining sources evaluate on the ImageNet dataset. The input size used in this benchmark is $224\times224\times3$.

\begin{table}
    \centering
    \caption{Benchmark results for image classification task on mid-range devices}
    \label{tab:benchmark_classification_midrange}
    \resizebox{0.48\textwidth}{!}{
    \begin{tabular}{llcrr}
        \toprule
        \textbf{Device} & \textbf{Model} & \textbf{Acc@1 (\%)} & \textbf{FPS} & \textbf{Latency (ms)} \\
        \midrule
        \multirow{3}{*}{\begin{tabular}{@{}l@{}}Jetson Nano\\\cite{Tobiasz2023EdgeDI}\end{tabular}}
        & MobileNetV2 & -    & 82.8   & 12.1 \\
        & InceptionV3 & -    & 36.4   & 27.5 \\
        & ResNet50    & -    & 35.1   & 28.5 \\
        \midrule
        \multirow{3}{*}{\begin{tabular}{@{}l@{}}Coral Dev Board\\\cite{coral_edgetpu_benchmarks}\end{tabular}}
        & MobileNetV1 & -    & 416.7  & 2.4 \\
        & MobileNetV2 & -    & 384.6  & 2.6 \\
        & InceptionV1 & -    & 243.9  & 4.1 \\
        \midrule
        \multirow{5}{*}{\begin{tabular}{@{}l@{}}Coral USB\\\cite{coral_image_classification}\end{tabular}}
        & MobileNetV1 & 69.5 & 357.1  & 2.8  \\
        & MobileNetV2 & 73.2 & 333.3  & 3.0  \\
        & InceptionV1 & 71.9 & 294.1  & 3.4  \\
        & InceptionV3 & 75.4 & 74.6   & 13.4 \\
        & ResNet50    & 73.6 & 23.7   & 42.2 \\
        \midrule
        \multirow{4}{*}{\begin{tabular}{@{}l@{}}Hailo Accelerator\\\cite{hailo_model_explorer}\end{tabular}}
        & MobileNetV1 & 70.3 & 3305.0 & 0.3 \\
        & MobileNetV2 & 70.9 & 2597.0 & 0.4 \\
        & InceptionV1 & 69.5 & 2454.0 & 0.4 \\
        & ResNet50    & 74.7 & 1372.0 & 0.7 \\
        \bottomrule
    \end{tabular}
    }
\end{table}

Overall, the benchmark results in Table~\ref{tab:benchmark_classification_midrange} demonstrate that lightweight architectures based on depthwise separable convolutions (such as MobileNetV1 and MobileNetV2) offer the best trade-off between latency and accuracy across mid-range hardware. These models comfortably exceed real-time requirements ($>80$\,FPS on embedded GPUs and $>300$\,FPS on dedicated accelerators) while maintaining competitive accuracy. In contrast, scaling to heavier backbones such as InceptionV3 or ResNet50 yields sharp diminishing returns, offering only marginal accuracy improvements ($+0.4\text{--}4.0\%$) at the expense of severe latency penalties---dropping throughput by up to $15\times$ on Edge TPU. Across hardware platforms, dedicated accelerators (such as Coral Edge TPU and Hailo-8) provide significantly higher frame rates for standard convolution layers than entry-level embedded GPUs, establishing lightweight CNNs as the standard baseline for mid-range edge devices.

Benchmark data for image classification task on microcontrollers in Table~\ref{tab:benchmark_classification_mcu} is compiled from two academic works~\cite{lin2023tiny, Sadiq2022TinyOpsIS}. Accuracy was evaluated on the ImageNet dataset.

\begin{table}
    \centering
    \caption{Benchmark results for image classification task on microcontrollers}
    \label{tab:benchmark_classification_mcu}
    \resizebox{0.48\textwidth}{!}{
    \begin{tabular}{lccrrr}
        \toprule
        \multirow{2}{*}{\textbf{Model}} & \multirow{2}{*}{\textbf{Res.}}
        & \textbf{Acc@1} & \textbf{Latency} & \textbf{Memory} & \textbf{Storage} \\
        & & \textbf{(\%)} & \textbf{(ms)} & \textbf{(KB)} & \textbf{(KB)} \\
        \midrule
        \multicolumn{6}{c}{\textbf{Device}: STM32H743; \textbf{Inference library}: TinyEngine~\cite{lin2020mcunet} (Source:~\cite{lin2023tiny})} \\
        \midrule
        MCUNet-10fps     & 48  & 40.4 & 25   & 49  & 842 \\
        MCUNet-5fps      & 96  & 49.9 & 56   & 96  & 727 \\
        MCUNet-256kb-1mb & 160 & 60.3 & 280  & 215 & 830 \\
        MCUNet-320kb-1mb & 176 & 61.8 & 336  & 260 & 835 \\
        MCUNet-512kb-2mb & 160 & 68.0 & 463  & 416 & 1825 \\
        Proxyless-w0.3   & 64  & 37.0 & 23   & 35  & 777 \\
        Proxyless-w0.3   & 176 & 56.2 & 176  & 259 & 779 \\
        MobileNetV2-w0.3 & 64  & 34.1 & 23   & 61  & 690 \\
        \midrule
        \multicolumn{6}{c}{\textbf{Device}: STM32H743; \textbf{Inference library}: TFLM~\cite{david2021tensorflow} (Source:~\cite{lin2023tiny})} \\
        \midrule
        MCUNet-10fps     & 48  & 40.4 & 596  & 161 & 1072 \\
        MCUNet-5fps      & 96  & 49.9 & 1227 & 219 & 937 \\
        MCUNet-256kb-1mb & 160 & 60.3 & 6463 & 460 & 1084 \\
        MCUNet-320kb-1mb & 176 & 61.8 & 7821 & 493 & 1091 \\
        MCUNet-512kb-2mb & 160 & 68.0 & OOM  & OOM & OOM \\
        Proxyless-w0.3   & 64  & 37.0 & 512  & 128 & 1065 \\
        Proxyless-w0.3   & 176 & 56.2 & 3801 & 453 & 1065 \\
        MobileNetV2-w0.3 & 64  & 34.1 & 467  & 173 & 940 \\
        \midrule
        \multicolumn{6}{c}{\textbf{Device}: STM32F469; \textbf{Inference library}: TinyOps (Source:~\cite{Sadiq2022TinyOpsIS})} \\
        \midrule
        Proxyless-w0.4   & 80  & 46.6 & 628  & 171 & 1300 \\
        MobileNetV3-w0.4 & 96  & 46.1 & 512  & 217 & 1157 \\
        MnasNet-w1.0     & 80  & 60.8 & 2575 & 330 & 4833 \\
        MnasNet-w1.0     & 96  & 64.0 & 2922 & 397 & 4833 \\
        \midrule
        \multicolumn{6}{c}{\textbf{Device}: STM32F746; \textbf{Inference library}: TinyOps (Source:~\cite{Sadiq2022TinyOpsIS})} \\
        \midrule
        Proxyless-w0.85  & 64  & 50.5 & 410  & 303 & 3564 \\
        MnasNet-w1.0     & 96  & 64.0 & 866  & 397 & 4833 \\
        MnasNet-w1.0     & 128 & 68.2 & 1423 & 568 & 4833 \\
        \bottomrule
    \end{tabular}
    }
\end{table}

At the microcontroller layer, the benchmark data in Table~\ref{tab:benchmark_classification_mcu} shows that inference performance is jointly determined by memory limits, model design, and the software inference library. In particular, the inference library plays a critical role; optimized runtimes like TinyEngine~\cite{lin2020mcunet} eliminate interpreter overhead and optimize memory reuse, achieving $10\times\text{--}20\times$ lower latency and avoiding out-of-memory (OOM) errors that occur under standard TFLM~\cite{david2021tensorflow}. Furthermore, because internal SRAM ($\le 512$\,KB) and Flash ($\le 2$\,MB) impose strict physical limits on MCU platforms, input resolution (from $48\times48$ to $176\times176$) is the main factor used to balance accuracy against SRAM usage, since activation memory grows quadratically with image dimensions. Hardware-aware models designed with neural architecture search (such as MCUNet) consistently outperform simple downscaled mobile backbones (such as MobileNetV2-w0.3), delivering much higher accuracy within tight sub-256\,KB SRAM budgets. Meanwhile, frameworks like TinyOps~\cite{Sadiq2022TinyOpsIS} help run wider mobile architectures on MCUs with larger external memory, expanding model choices when memory constraints are less restrictive.

\subsection{Object detection}

Benchmark data for object detection task on mid-range devices in Table~\ref{tab:benchmark_detection_midrange} is compiled from Alqahtani et al.~\cite{Alqahtani2024ACE}, Coral Models for object detection~\cite{coral_object_detection}, and Hailo AI Model Explorer for computer vision~\cite{hailo_model_explorer}. The models are evaluated on the COCO dataset with mAP@[0.5:0.95].

\begin{table}
    \centering
    \caption{Benchmark results for object detection task on mid-range devices}
    \label{tab:benchmark_detection_midrange}
    \resizebox{0.48\textwidth}{!}{
    \begin{tabular}{llccrr}
        \toprule
        \textbf{Device} & \textbf{Model} & \textbf{Res.} & \textbf{mAP (\%)} & \textbf{FPS} & \textbf{Latency (ms)} \\
        \midrule
        \multirow{8}{*}{\begin{tabular}{@{}l@{}}RPi5 + Coral USB\\\cite{Alqahtani2024ACE}\end{tabular}}
        & SSD-MobileNetV1    & 300 & 19.0 & 100.0  & 10.0 \\
        & SSDLite-MobileDet  & 320 & 30.0 & 66.7   & 15.0 \\
        & EfficientDet-Lite0 & 320 & 26.0 & 17.5   & 57.0 \\
        & EfficientDet-Lite1 & 384 & 30.0 & 11.9   & 84.0 \\
        & EfficientDet-Lite2 & 448 & 33.0 & 7.2    & 139.0 \\
        & YOLOv8n            & 320 & 16.0 & 47.6   & 21.0 \\
        & YOLOv8s            & 320 & 25.0 & 27.8   & 36.0 \\
        & YOLOv8m            & 320 & 32.0 & 10.3   & 97.0 \\
        \midrule
        \multirow{8}{*}{\begin{tabular}{@{}l@{}}RPi5 + AI HAT+\\\cite{Alqahtani2024ACE}\end{tabular}}
        & SSD-MobileNetV1    & 300 & 17.0 & 11.2   & 89.0 \\
        & SSDLite-MobileDet  & 320 & 27.0 & 10.9   & 92.0 \\
        & EfficientDet-Lite0 & 320 & 17.0 & 8.8    & 113.0 \\
        & EfficientDet-Lite1 & 384 & 22.0 & 7.8    & 128.0 \\
        & EfficientDet-Lite2 & 448 & 24.0 & 6.8    & 147.0 \\
        & YOLOv8n            & 640 & 32.0 & 13.2   & 76.0 \\
        & YOLOv8s            & 640 & 40.0 & 6.9    & 144.0 \\
        & YOLOv8m            & 640 & 44.0 & 4.2    & 240.0 \\
        \midrule
        \multirow{8}{*}{\begin{tabular}{@{}l@{}}Jetson Nano\\\cite{Alqahtani2024ACE}\end{tabular}}
        & SSD-MobileNetV1    & 300 & 16.0 & 19.2   & 52.0 \\
        & SSDLite-MobileDet  & 320 & 27.0 & 15.6   & 64.0 \\
        & EfficientDet-Lite0 & 320 & 23.0 & 20.8   & 48.0 \\
        & EfficientDet-Lite1 & 384 & 28.0 & 11.5   & 87.0 \\
        & EfficientDet-Lite2 & 448 & 32.0 & 7.8    & 129.0 \\
        & YOLOv8n            & 640 & 16.0 & 9.6    & 104.0 \\
        & YOLOv8s            & 640 & 25.0 & 4.4    & 226.0 \\
        & YOLOv8m            & 640 & 32.0 & 2.0    & 509.0 \\
        \midrule
        \multirow{8}{*}{\begin{tabular}{@{}l@{}}Jetson Orin Nano\\\cite{Alqahtani2024ACE}\end{tabular}}
        & SSD-MobileNetV1    & 300 & 16.0 & 45.5   & 22.0 \\
        & SSDLite-MobileDet  & 320 & 27.0 & 52.6   & 19.0 \\
        & EfficientDet-Lite0 & 320 & 23.0 & 50.0   & 20.0 \\
        & EfficientDet-Lite1 & 384 & 28.0 & 43.5   & 23.0 \\
        & EfficientDet-Lite2 & 448 & 32.0 & 34.5   & 29.0 \\
        & YOLOv8n            & 640 & 16.0 & 62.5   & 16.0 \\
        & YOLOv8s            & 640 & 25.0 & 45.5   & 22.0 \\
        & YOLOv8m            & 640 & 32.0 & 20.0   & 50.0 \\
        \midrule
        \multirow{7}{*}{\begin{tabular}{@{}l@{}}Coral USB\\\cite{coral_object_detection}\end{tabular}}
        & SSD-MobileNetV1    & 300 & 21.5 & 153.8  & 6.5 \\
        & SSD-MobileNetV2    & 300 & 25.6 & 137.0  & 7.3 \\
        & SSDLite-MobileDet  & 320 & 32.9 & 109.9  & 9.1 \\
        & EfficientDet-Lite0 & 320 & 30.4 & 26.7   & 37.4 \\
        & EfficientDet-Lite1 & 384 & 34.3 & 17.8   & 56.3 \\
        & EfficientDet-Lite2 & 448 & 36.0 & 9.6    & 104.6 \\
        & EfficientDet-Lite3 & 512 & 39.4 & 9.3    & 107.6 \\
        \midrule
        \multirow{20}{*}{\begin{tabular}{@{}l@{}}Hailo Accelerator\\\cite{hailo_model_explorer}\end{tabular}}
        & SSD-MobileNetV1    & 300 & 22.4 & 1015.0 & 1.0 \\
        & SSD-MobileNetV2    & 300 & 23.0 & 784.0  & 1.3 \\
        & EfficientDet-Lite0 & 320 & 26.5 & 109.0  & 9.2 \\
        & EfficientDet-Lite1 & 384 & 31.8 & 76.8   & 13.0 \\
        & EfficientDet-Lite2 & 448 & 34.7 & 50.2   & 19.9 \\
        & YOLOv5s            & 640 & 34.1 & 543.0  & 1.8 \\
        & YOLOv5m            & 640 & 41.3 & 156.0  & 6.4 \\
        & YOLOv8n            & 640 & 36.4 & 1036.0 & 1.0 \\
        & YOLOv8s            & 640 & 44.0 & 491.0  & 2.0 \\
        & YOLOv8m            & 640 & 49.2 & 67.3   & 14.9 \\
        & YOLOv8l            & 640 & 51.8 & 36.3   & 27.5 \\
        & YOLOv8x            & 640 & 52.9 & 18.0   & 55.6 \\
        & YOLOv11n           & 640 & 37.8 & 185.0  & 5.4 \\
        & YOLOv11s           & 640 & 45.5 & 111.0  & 9.0 \\
        & YOLOv11m           & 640 & 49.8 & 50.2   & 19.9 \\
        & YOLOv11l           & 640 & 52.0 & 38.0   & 26.3 \\
        & YOLOv11x           & 640 & 53.1 & 14.2   & 70.4 \\
        & YOLO26n            & 640 & 37.4 & 161.0  & 6.2 \\
        & YOLO26s            & 640 & 44.8 & 97.6   & 10.2 \\
        & YOLO26m            & 640 & 50.0 & 48.6   & 20.6 \\
        \bottomrule
    \end{tabular}
    }
\end{table}

The mid-range object detection benchmarks in Table~\ref{tab:benchmark_detection_midrange} highlight how detector architectures, hardware accelerators, and host integration interact. First, hardware compatibility varies significantly across model families: Edge TPU accelerators achieve high frame rates with SSD and MobileDet architectures ($>100$\,FPS) but slow down considerably on models with complex feature networks like EfficientDet. In contrast, Hailo accelerators excel at modern YOLO architectures (YOLOv8/v11), while embedded GPUs (Jetson Orin Nano) offer broad architectural flexibility, delivering consistent performance across both SSD and YOLO models. Second, input resolution is the most critical factor for detection accuracy; moving from $320\times320$ to $640\times640$ inputs substantially improves mAP, making compact YOLO variants (such as YOLOv8n/s) an effective choice for balancing accuracy and real-time speed. Finally, as analyzed in Section~\ref{sec:hardware}, when discrete accelerators are paired with single-board computers (such as Raspberry Pi 5 with Coral USB or AI HAT+), host CPU processing (e.g., Non-Maximum Suppression) and communication over USB or PCIe become the main throughput bottlenecks when raw accelerator inference takes only a few milliseconds.

Benchmark data for object detection on microcontrollers in Table~\ref{tab:benchmark_detection_mcu} is compiled from~\cite{Sah2024MCUBenchAB}. The models are evaluated on the Pascal VOC dataset with mAP@0.5.

\begin{table}
    \centering
    \caption{Benchmark results for object detection task on microcontrollers (Source: Sah et al.~\cite{Sah2024MCUBenchAB})}
    \label{tab:benchmark_detection_mcu}
    \resizebox{0.48\textwidth}{!}{
    \begin{tabular}{llccccr}
        \toprule
        \textbf{Device} & \textbf{Model} & \textbf{Res.} & \textbf{Depth} & \textbf{Width} & \textbf{mAP} & \textbf{Latency (ms)} \\
        \midrule
        \multirow{2}{*}{NUCLEO-H743ZI}
        & YOLOv6 & 128 & 0.085 & 0.050 & 0.08 &  100 \\
        & YOLOv7 & 224 & 0.250 & 0.160 & 0.35 & 1300 \\
        \midrule
        \multirow{2}{*}{B-U585I-IOT02A}
        & YOLOv6 & 128 & 0.085 & 0.050 & 0.07 &  400 \\
        & YOLOv7 & 224 & 0.250 & 0.250 & 0.39 & 8800 \\
        \midrule
        \multirow{2}{*}{STM32F469I-DISCO}
        & YOLOv3 & 128 & 0.160 & 0.125 & 0.17 & 1000 \\
        & YOLOv3 & 224 & 0.250 & 0.250 & 0.39 & 9230 \\
        \midrule
        \multirow{2}{*}{STM32F769I-DISCO}
        & YOLOv6 & 128 & 0.085 & 0.050 & 0.08 &  220 \\
        & YOLOv7 & 224 & 0.250 & 0.250 & 0.39 & 4370 \\
        \midrule
        \multirow{2}{*}{STM32H573I-DK}
        & YOLOv6 & 128 & 0.085 & 0.050 & 0.07 &  260 \\
        & YOLOv7 & 224 & 0.250 & 0.250 & 0.39 & 5040 \\
        \midrule
        \multirow{2}{*}{STM32H747I-DISCO}
        & YOLOv6 & 128 & 0.085 & 0.050 & 0.08 &  120 \\
        & YOLOv7 & 224 & 0.250 & 0.200 & 0.41 & 2210 \\
        \midrule
        \multirow{2}{*}{STM32L4R9I-DISCO}
        & YOLOv6 & 128 & 0.085 & 0.050 & 0.07 &  660 \\
        & YOLOv7 & 224 & 0.250 & 0.250 & 0.39 & 12800 \\
        \bottomrule
    \end{tabular}
    }
\end{table}

At the microcontroller layer, the benchmark data in Table~\ref{tab:benchmark_detection_mcu} illustrates the practical performance boundaries of TinyML object detection across different MCU platforms. As compiled from~\cite{Sah2024MCUBenchAB}, the table presents the two boundary configurations for each device: a minimum-latency setup (for fast detection) and a maximum-accuracy setup. To fit within the strict SRAM and Flash constraints of MCUs, YOLO models must be scaled down heavily in depth and width (down to $5\text{--}25\%$ of the original size). The minimum-latency configurations (typically YOLOv6 at $128\times128$ resolution) achieve fast response times ($100\text{--}660$\,ms) suitable for quick triggering, but only reach modest detection accuracy ($7\text{--}17\%$ mAP). Conversely, the maximum-accuracy configurations (YOLOv7 or YOLOv3 at $224\times224$) increase accuracy to $35\text{--}41\%$ mAP, but at the cost of high latencies ranging from $1.3$\,s to nearly $13$\,s. Because these MCUs lack hardware accelerators and execute all computations sequentially on general-purpose CPUs, processing speed is heavily determined by CPU core architecture and clock frequency: high-performance Cortex-M7 cores ($216\text{--}480$\,MHz) can achieve sub-second response times for small models, whereas lower-power Cortex-M4 and Cortex-M33 cores ($120\text{--}180$\,MHz) require several seconds per frame.

\section{Hardware-to-Model Mapping Framework}\label{sec:framework}
Synthesizing empirical benchmark results (Level~4, Section~\ref{sec:benchmark}) with hardware constraints (Level~1, Section~\ref{sec:hardware}), model architectures (Level~2, Section~\ref{sec:model}), and deployment software runtimes (Level~3, Section~\ref{sec:software}), this section formalizes \textbf{Level~5 (Actionable Design Decisions)} of the deployment framework (Fig.~\ref{fig:framework_overview}). It addresses the second research question (\textbf{RQ2}) across three foundational decision dimensions: (1) \textit{Model Selection Guidelines}, mapping hardware tiers and priority regimes (Speed, Balanced, Accuracy) to recommended off-the-shelf models; (2) \textit{Architectural Customization Principles}, providing SRAM-aware topology scaling rules; and (3) \textit{Runtime Optimization Principles}, guiding algorithm--engine co-design and compiler quantization.

\subsection{Off-the-Shelf Model Selection Guidelines}

For existing deployments, model selection should follow a hardware-first workflow based on the available accelerator and application priority (Speed, Balanced, or Accuracy).

\subsubsection{Mid-range devices}
On mid-range platforms, selection is governed by the compatibility between the model topology and the accelerator type. Tables~\ref{tab:discussion_midrange_cls} and~\ref{tab:discussion_midrange_det} provide concrete recommendations across three priority tiers. On Edge TPUs (Coral), depthwise-separable CNNs (MobileNetV1/V2) and MobileDet detectors achieve maximum throughput, whereas complex multi-scale feature fusion networks (EfficientDet) suffer significant latency penalties. On dataflow NPUs (Hailo-8), modern YOLO variants (YOLOv8/v11) offer superior accuracy and frame rates. On embedded GPUs (Jetson Orin Nano), the broad operator flexibility allows deploying either YOLO or EfficientDet models depending on whether frame rate or multi-scale detection is prioritized. For discrete accelerators connected to SBCs (e.g., Raspberry Pi 5), actual end-to-end throughput will be lower than standalone server benchmarks due to host CPU pre/post-processing and bus communication overhead.

\begin{table}
    \centering
    \caption{Model recommendations for image classification task on mid-range devices}
    \label{tab:discussion_midrange_cls}
    \resizebox{0.48\textwidth}{!}{
    \begin{tabular}{llcrr}
        \toprule
        \textbf{Accelerator} & \textbf{Priority} & \textbf{Model} & \textbf{Acc@1 (\%)} & \textbf{FPS} \\
        \midrule
        \multirow{3}{*}{Edge TPU (Coral USB)}
        & Speed     & MobileNetV1 & 69.5 & 357  \\
        & Balanced  & MobileNetV2 & 73.2 & 333  \\
        & Accuracy  & InceptionV3 & 75.4 & 75   \\
        \midrule
        \multirow{3}{*}{Hailo-8}
        & Speed     & MobileNetV1 & 70.3 & 3305 \\
        & Balanced  & MobileNetV2 & 70.9 & 2597 \\
        & Accuracy  & ResNet50    & 74.7 & 1372 \\
        \midrule
        GPU (Jetson Nano)
        & Speed     & MobileNetV2 & --   & 82.8 \\
        \bottomrule
        \multicolumn{5}{l}{\footnotesize Jetson Nano: only FPS data available, no acc@1.} \\
    \end{tabular}
    }
\end{table}

\begin{table}
    \centering
    \caption{Model recommendations for object detection task on mid-range devices}
    \label{tab:discussion_midrange_det}
    \resizebox{0.48\textwidth}{!}{
    \begin{tabular}{llcrr}
        \toprule
        \textbf{Accelerator} & \textbf{Priority} & \textbf{Model} & \textbf{mAP (\%)} & \textbf{FPS} \\
        \midrule
        \multirow{3}{*}{Edge TPU (Coral USB)}
        & Speed     & SSD-MobileNetV1    & 21.5 & 153.8 \\
        & Balanced  & SSDLite-MobileDet  & 32.9 & 109.9 \\
        & Accuracy  & EfficientDet-Lite3 & 39.4 & 9.3   \\
        \midrule
        \multirow{3}{*}{Hailo-8}
        & Speed     & YOLOv8n            & 36.4 & 1036  \\
        & Balanced  & YOLOv8s            & 44.0 & 491   \\
        & Accuracy  & YOLOv11x           & 53.1 & 14.2  \\
        \midrule
        \multirow{3}{*}{GPU (Jetson Orin Nano)}
        & Speed     & YOLOv8n            & 16.0 & 62.5  \\
        & Balanced  & SSDLite-MobileDet  & 27.0 & 52.6  \\
        & Accuracy  & EfficientDet-Lite2 & 32.0 & 34.5  \\
        \bottomrule
    \end{tabular}
    }
\end{table}

\subsubsection{Microcontrollers (MCUs)}
On microcontrollers, internal SRAM is the primary hard boundary. Table~\ref{tab:discussion_mcu} maps the SRAM budget to recommended image classification models on an STM32H743 using TinyEngine. For devices with $\le 256$\,KB SRAM, specialized NAS models such as MCUNet at lower input resolutions ($48\text{--}160$\,px) provide the only viable route to achieve acceptable accuracy without OOM. In the $256\text{--}512$\,KB range, higher-resolution MCUNet variants or ProxylessNAS offer balanced-to-high accuracy ($60\text{--}68\%$).

For MCU object detection, model selection depends on the target operational regime as summarized in Table~\ref{tab:discussion_mcu_det}. Developers requiring fast wake-up or coarse triggering ($100\text{--}660$\,ms latency) should select heavily downscaled YOLO models ($d \approx 0.08, w \approx 0.05$) at $128\times128$ resolution. Conversely, applications requiring higher localization fidelity ($35\text{--}41\%$ mAP) must allocate a multi-second latency budget ($1.3\text{--}12.8$\,s) with $224\times224$ models, which operate with practical sub-2-second latency only on high-performance Cortex-M7 platforms.

\begin{table}
    \centering
    \caption{SRAM budget -- model mapping for image classification task on microcontrollers (STM32H743, TinyEngine)}
    \label{tab:discussion_mcu}
    \resizebox{0.48\textwidth}{!}{
    \begin{tabular}{llllrr}
        \toprule
        \textbf{SRAM} & \textbf{Priority} & \textbf{Model} & \textbf{Res.} & \textbf{Acc@1 (\%)} & \textbf{Latency (ms)} \\
        \midrule
        \multirow{3}{*}{$\leq$ 256KB}
        & Speed     & MCUNet-10fps       & 48  & 40.4 & 25  \\
        & Balanced  & MCUNet-5fps        & 96  & 49.9 & 56  \\
        & Accuracy  & MCUNet-256kb-1mb   & 160 & 60.3 & 280 \\
        \midrule
        \multirow{3}{*}{256--512KB}
        & Speed     & Proxyless-w0.3     & 176 & 56.2 & 176 \\
        & Balanced  & MCUNet-320kb-1mb   & 176 & 61.8 & 336 \\
        & Accuracy  & MCUNet-512kb-2mb   & 160 & 68.0 & 463 \\
        \bottomrule
    \end{tabular}
    }
\end{table}

\begin{table}
    \centering
    \caption{Model recommendations for object detection task on microcontrollers}
    \label{tab:discussion_mcu_det}
    \resizebox{0.48\textwidth}{!}{
    \begin{tabular}{lllcrr}
        \toprule
        \textbf{Target Core} & \textbf{Priority / Regime} & \textbf{Model Architecture} & \textbf{Res.} & \textbf{mAP (\%)} & \textbf{Latency (ms)} \\
        \midrule
        \multirow{2}{*}{Cortex-M7 (High-perf.)}
        & Fast Trigger  & YOLOv6 ($d=0.08, w=0.05$) & 128 & 8.0 & 100--120 \\
        & High Accuracy & YOLOv7 ($d=0.25, w=0.16\text{--}0.20$) & 224 & 35.0--41.0 & 1300--2210 \\
        \midrule
        \multirow{2}{*}{Cortex-M4/M33 (Low-power)}
        & Fast Trigger  & YOLOv6 ($d=0.08, w=0.05$) & 128 & 7.0--8.0 & 260--660 \\
        & High Accuracy & YOLOv7 ($d=0.25, w=0.25$) & 224 & 39.0 & 4370--12800 \\
        \bottomrule
    \end{tabular}
    }
\end{table}

\subsection{Architectural Customization and Runtime Optimization Principles}

When adapting existing vision networks or designing new architectures from scratch, practitioners should apply the following hardware-aware principles derived from our architectural and benchmark analysis:

\subsubsection{Mid-Range Hardware (Accelerators and GPUs)}
For edge platforms equipped with dedicated NPUs or embedded GPUs, model design should balance operator support, feature resolution, and host communication:

\textbf{Operator and Topology Alignment:} Dedicated accelerators (Edge TPU, Hailo-8) rely on efficient dataflow through on-chip memory. Designers should prioritize standard $1\times1$ and $3\times3$ convolutions, depthwise separable blocks, and linear bottlenecks. Complex non-sequential branches, dynamic routing, and multi-scale bidirectional connections (e.g., BiFPN) should be avoided on dedicated NPUs because they cause frequent off-chip memory access, whereas embedded GPUs handle such branching efficiently.

\textbf{Resolution-First Detector Scaling:} For object detection, benchmark data confirms that increasing input resolution (e.g., $640\times640$) with a lightweight backbone (e.g., YOLOv8n) yields significantly higher mAP than running a heavier backbone (e.g., YOLOv8m) at reduced resolution ($320\times320$), while preserving higher frame rates.

\textbf{Host-Interconnect Overhead Mitigation:} On systems where discrete accelerators are hosted by an SBC (e.g., Raspberry Pi), design choices should minimize host-side workloads. Anchor-free detection heads and lightweight post-processing layers reduce host CPU bottleneck during Non-Maximum Suppression (NMS).

\subsubsection{Microcontrollers (TinyML)}
For bare-metal MCUs constrained by sub-megabyte memory and pure CPU execution, model design must strictly adhere to the following principles:

\textbf{SRAM-Centric Peak Memory Budgeting:} On MCUs, the primary architectural design constraint is peak SRAM activation memory, not total parameter count. Because activation memory scales quadratically with spatial input dimensions, architects must downscale early feature map dimensions quickly using aggressive striding or inverted residual bottlenecks, ensuring that the largest activation layer fits within internal SRAM ($<256\text{--}512$\,KB).

\textbf{Algorithm--System Co-Design with Dedicated Runtimes:} Model architecture search must be coupled directly with the inference engine (e.g., TinyEngine). Hand-crafted or uniformly scaled mobile networks (e.g., MobileNetV2-w0.3) waste memory on intermediate buffers; co-designed NAS models that optimize operator scheduling and in-place buffer reuse achieve substantially higher accuracy under identical hardware budgets.

\textbf{Hardware-Friendly Operators and Quantization Uniformity:} Models targeting pure Cortex-M CPUs must strictly adhere to 8-bit integer (INT8) quantization across all layers to utilize ARM DSP/SIMD instruction sets (CMSIS-NN). Architects should avoid complex non-linear operations (such as non-linear activations in bottleneck layers, complex attention maps, or dynamic tensors) in favor of standard ReLU/ReLU6 and static memory graphs.

\section{Challenges and Future Directions}\label{sec:challenge}
While edge computer vision has progressed rapidly through lightweight architectures and hardware-aware neural architecture search, deploying deep learning models on constrained, heterogeneous edge hardware remains bottlenecked by critical ecosystem, compiler, and algorithmic challenges. This section outlines six core deployment challenges and identifies three promising future research directions.

\subsection{Core Deployment Challenges}

\textbf{Software and Compiler Toolchain Fragmentation:} Achieving high performance on edge hardware requires compiling models into vendor-specific low-level instructions. However, the software ecosystem remains severely fragmented. Mid-range accelerators rely on proprietary runtimes and compilers (such as NVIDIA TensorRT, Google Edge TPU Compiler, and HailoRT)~\cite{baller2021deepedgebench, ratul2025accelerating, Tobiasz2023EdgeDI}, each requiring dedicated calibration and quantization pipelines. At the microcontroller layer and for emerging dedicated MPU/SoCs (e.g., STM32N6, GAP9, and Himax WiseEye2), developers must navigate isolated vendor toolchains (ST X-Cube-AI, GreenWaves GAP SDK)~\cite{carboni2026tiny, muller2024gap9shield, millar2025benchmarking}. The absence of a universal intermediate representation (IR) or portable runtime prevents seamless cross-platform deployment and increases engineering overhead.

\textbf{Benchmark Fragmentation and Standardization:} Evaluating and comparing edge vision models across the literature is notoriously difficult due to inconsistent experimental setups. As documented in Section~\ref{sec:benchmark}, benchmark studies vary widely in host CPU capabilities, compiler optimization flags, input resolutions, and datasets (ImageNet, COCO, Pascal VOC)~\cite{banbury2021mlperf, Sah2024MCUBenchAB, Sadiq2022TinyOpsIS}. Furthermore, power measurements are heavily influenced by board-level peripherals, GPIO configurations, and cooling solutions. While initiatives like MLPerf Tiny~\cite{banbury2021mlperf} provide standardized benchmarking harnesses, their scope has historically focused on simple, non-dense tasks (e.g., Visual Wake Words), leaving a standard benchmarking gap for complex object detection and segmentation tasks.

\textbf{Operator and Hardware Incompatibility:} Hardware accelerators (NPUs, TPUs, and DSPs) achieve extreme energy efficiency by hardcoding specific matrix-multiplication dataflows, sacrificing general-purpose computing versatility. When an edge compiler encounters an unsupported operator (such as custom pooling layers, dynamic tensor operations, or non-linear activations), it silently offloads execution back to the host CPU. This CPU fallback causes severe latency and energy penalties due to memory marshaling and bus communication~\cite{baller2021deepedgebench, ratul2025accelerating}. On low-power MCUs, advanced activation functions (such as the hard-swish in MobileNetV3~\cite{Howard2019SearchingFM}) often lack low-overhead assembly kernels in standard runtimes like TFLM~\cite{david2021tensorflow}, forcing developers to trade accuracy for standard ReLU or endure unoptimized software emulation.

\textbf{Vision Transformer (ViT) Feasibility Boundaries:} Although Vision Transformers offer high representational capacity, their deployment on constrained edge devices remains severely limited~\cite{papa2024survey, Saha2025VisionTO}. Standard self-attention scales quadratically ($O(N^2)$) with token length, imposing severe computational and memory bandwidth bottlenecks. Moreover, edge accelerators frequently lack optimized hardware kernels for LayerNorm and GELU, necessitating costly dequantization back to 32-bit floating-point (FP32) during execution. On microcontrollers with sub-megabyte SRAM, standard ViTs remain infeasible; practitioners must rely on hybrid architectures (such as MobileViT~\cite{mehta2021mobilevit}, EfficientFormer~\cite{li2022efficientformer}, or FastViT~\cite{vasu2023fastvit}) that restrict self-attention to late network stages.

\textbf{Narrow Task Scope and Peak Activation Memory Wall:} The majority of edge model compression and NAS research is heavily biased toward image classification, whereas dense prediction tasks (object detection and semantic segmentation) present fundamentally different memory bottlenecks~\cite{Lin2021MCUNetV2MP, Sah2024MCUBenchAB}. While parameter weights can be compressed and stored in external Flash memory, intermediate activation tensors must reside entirely in high-speed on-chip SRAM during execution. Because dense prediction tasks require high-resolution input grids to resolve spatial details, peak activation memory in early feature layers frequently exceeds the physical SRAM budget ($<256\text{--}512$\,KB), causing immediate out-of-memory crashes regardless of parameter compactness.

\textbf{The Research-to-Practice Translation Gap:} Advanced optimization methods widely studied in academia face significant practical friction on real edge devices. For example, unstructured weight pruning produces irregular, non-contiguous sparsity patterns that commodity neural accelerators and SIMD instruction sets cannot accelerate efficiently, yielding negligible physical speedup~\cite{lin2023tiny, Somvanshi2025FromTM}. Furthermore, complex retraining-heavy compression pipelines (such as multi-stage knowledge distillation) require deep cross-disciplinary expertise across hardware, compilers, and machine learning, leading industrial practitioners to favor simpler post-training INT8 quantization and resolution scaling.

\subsection{Future Research Directions}

\textbf{On-Device Training and Continual Adaptation:} Transitioning edge devices from static inference engines to dynamic learning agents is essential to mitigate real-world concept drift (e.g., shifting environmental conditions in outdoor surveillance) while preserving user privacy~\cite{Somvanshi2025FromTM, lin2023tiny}. However, backpropagation requires caching intermediate activations across all layers, increasing peak memory footprint by up to $6.9\times$ compared to inference~\cite{lin2022ondevice}. Promising research avenues include sparse-update training (updating only bias vectors or select bottleneck layers) and quantization-aware gradient scaling to support stable backward passes directly in low-bit integer formats without memory-intensive floating-point fallbacks.

\textbf{Unified Hardware--Software--Algorithm Co-Design:} Historically, model architecture design, compression, compiler optimization, and hardware design have operated as isolated, sequential stages. To maximize edge performance, future research must transition toward unified multi-objective co-design pipelines. Rather than optimizing for indirect proxies such as parameter count or FLOPs, neural architecture search loops must directly integrate on-device profiling metrics (physical latency, dynamic energy consumption, and peak SRAM footprints)~\cite{cai2018proxylessnas, lin2020mcunet}. Expanding co-design frameworks to jointly optimize hardware accelerator parameters (e.g., array dimensions, buffer sizes) alongside neural layer topologies represents a critical frontier.

\textbf{System-Level Sustainable Edge AI:} Traditional evaluation frameworks isolate individual model inference metrics. However, practical edge systems operate under strict whole-system energy and lifetime budgets. Future research must establish sustainable AI metrics that jointly evaluate sensor acquisition costs, on-chip computation, wireless transmission overhead, and dynamic duty-cycling. Developing adaptive, input-dependent elastic models that dynamically modulate network depth, width, and resolution based on real-time battery state-of-charge or ambient energy-harvesting rates is vital for self-sustaining autonomous edge intelligence.

\section{Conclusion}\label{sec:conclusion}
This paper has presented a systematic study bridging edge hardware constraints, vision model architectures, and deployment software ecosystems. We classified edge hardware into three distinct tiers—mid-range devices, dedicated embedded MPU/SoCs, and microcontrollers—and analyzed their core bottlenecks across computation, memory, and energy. By synthesizing multi-source benchmark data for image classification and object detection, we demonstrated the empirical trade-offs between model topology, input resolution, and compiler runtimes across platforms.

Building upon these findings, we established a comprehensive hardware-to-model mapping framework that delivers actionable guidelines for both off-the-shelf model selection and hardware-aware architectural design (such as SRAM-centric activation budgeting, resolution-first scaling, and algorithm--system co-design). Finally, we examined remaining deployment bottlenecks—spanning toolchain fragmentation, benchmark inconsistencies, and operator support gaps—and highlighted vital future research frontiers in on-device continual learning, unified co-design pipelines, and sustainable system-level Edge AI.

\renewcommand{\refname}{Reference}
\bibliographystyle{IEEEtran}
\bibliography{main}

\end{document}